\pdfoutput=1

\documentclass[11pt, final]{article}

\usepackage[dvipsnames]{xcolor}
\usepackage[final]{acl}

\usepackage{times}
\usepackage{latexsym}

\usepackage[T1]{fontenc}
\usepackage{bold-extra}

\usepackage[utf8]{inputenc}

\usepackage{microtype}

\usepackage{inconsolata}

\usepackage{graphicx}

\usepackage{url}
\usepackage{amsmath,amsfonts,amssymb,amsthm,mathtools}

\usepackage{cleveref}

\usepackage{bm}
\usepackage{paralist}
\usepackage{changepage}

\usepackage{subcaption}
\usepackage{floatrow}

\usepackage{tabularx}
\usepackage{booktabs}
\usepackage{pdflscape}
\usepackage{multirow}
\usepackage{makecell}
\usepackage{afterpage}
\usepackage{dcolumn}
\usepackage{array}
\newcolumntype{L}[1]{>{\raggedright\let\newline\\\arraybackslash\hspace{0pt}}m{#1}}
\newcolumntype{R}[1]{>{\raggedleft\let\newline\\\arraybackslash\hspace{0pt}}m{#1}}

\usepackage{svg}

\usepackage{enumerate}
\usepackage{enumitem}
\providecommand{\tightlist}{\setlength{\itemsep}{0pt}\setlength{\parskip}{0pt}\setlength{\topsep}{0pt}}

\usepackage{soul}
\usepackage{color}

\definecolor{customblue}{RGB}{0, 84, 168}
\definecolor{promptBlue}{RGB}{186,225,255} 
\definecolor{promptGreen}{RGB}{186,255,201}   
\definecolor{promptRed}{RGB}{255,179,186}     

\definecolor{lgray}{RGB}{176, 179, 184}

\newcommand{\llamasmall}{\texttt{Llama-3.1-8B}}
\newcommand{\llamabig}{\texttt{Llama-3.3-70B}}
\newcommand{\qwensmall}{\texttt{Qwen-2.5-7B}}
\newcommand{\qwenmid}{\texttt{Qwen-2.5-32B}}
\newcommand{\qwenbig}{\texttt{Qwen-2.5-72B}}

\newcommand{\llamai}{\texttt{Llama-3.1-8B-Instruct}}
\newcommand{\deberta}{\texttt{DeBERTa-v3-large}}
\newcommand{\modbert}{\texttt{ModernBERT-large}}

\DeclareMathOperator*{\argmax}{argmax}

\usepackage{listings}
\usepackage[most]{tcolorbox}

\newtcolorbox[list inside=prompt,auto counter,number within=section]{prompt}[1][]{
    colbacktitle=black!60,
    fonttitle=\small,
    coltitle=white,
    fontupper=\footnotesize,
    boxsep=3pt,
    left=0pt,
    right=0pt,
    top=0pt,
    bottom=0pt,
    boxrule=1pt,
    #1,
    breakable,              %
}

\usepackage{bbm}
\usepackage{todonotes}

\makeatletter
\newcommand*\iftodonotes{
    \if@todonotes@disabled
        \expandafter\@secondoftwo
    \else
        \expandafter\@firstoftwo
    \fi
}  %
\makeatother

\newif\ifhidecomments
\hidecommentstrue %

\ifhidecomments
\fi

\newcommand{\ImmFear}{\textsc{\mbox{Immigration Fear}}}
\newcommand{\AdNeg}{\textsc{\mbox{Ad-Negativity}}}
\newcommand{\Grand}{\textsc{\mbox{Grandstanding}}}

\title{%
    Measuring Scalar Constructs in Social Science with LLMs
}

\usepackage{tipa}
\newcommand{\inns}{\boldsymbol{\texttt{1}}}
\newcommand{\mary}{\boldsymbol{\texttt{2}}}
\newcommand{\american}{\boldsymbol{\texttt{3}}}
\newcommand{\north}{\boldsymbol{\texttt{4}}}
\newcommand{\ethz}{\boldsymbol{\texttt{5}}}

\author{
Hauke Licht\thanks{Equal contribution}$^{\inns}$~\;~
Rupak Sarkar\footnotemark[1]$^{\mary}$~\;~
Patrick Y. Wu$^{\american}$~\;~
Pranav Goel$^{\north}$\\
\textbf{Niklas Stoehr$^{\ethz}$}~\;~
\textbf{Elliott Ash$^{\ethz}$}~\;~
\textbf{Alexander Miserlis Hoyle\footnotemark[1]}$^{\ethz}$~\;~ 
\\
$^{\inns}$University of Innsbruck \quad
$^{\mary}$University of Maryland \quad
$^{\american}$American University\\
$^{\north}$Northeastern University \quad
$^{\ethz}$ETH Z{\"u}rich
\\
\small
{\url{hauke.licht@uibk.ac.at}}~\;~
{\url{rupak@umd.edu}}~\;~ 
{\url{patrickwu@american.edu}}~\;~ 
{\url{p.goel@northeastern.edu}}\\ 
\small
{\url{niklas.stoehr@inf.ethz.ch}}~\;~ 
{\url{elliott.ash@gess.ethz.ch}}~\;~
{\url{alexander.hoyle@ai.ethz.ch}}
}

\begin{document}
\maketitle
\begin{abstract}

Many constructs that characterize language, like its complexity or emotionality, have a naturally continuous semantic structure; a public speech is not just ``simple'' or ``complex,'' but exists on a continuum between extremes.
Although large language models (LLMs) are an attractive tool for measuring scalar constructs, their idiosyncratic treatment of numerical outputs raises questions of how to best apply them. 
We address these questions with a comprehensive evaluation of LLM-based approaches to scalar construct measurement in social science.
Using multiple datasets sourced from the political science literature, we evaluate four approaches: unweighted direct pointwise scoring, aggregation of pairwise comparisons, token-probability-weighted pointwise scoring, and finetuning.
Our study finds that pairwise comparisons made by LLMs produce better measurements than simply prompting the LLM to 
directly output the scores, which suffers from bunching around arbitrary numbers. 
However, taking the weighted mean over the token probability of scores further improves the measurements over the two previous approaches. 
Finally, finetuning smaller models with as few as 1,000 training pairs can match or exceed the performance of prompted LLMs. 
\end{abstract}

\section{Introduction}\label{sec:introduction}

\begin{figure}[!th]
    \centering
    \begin{adjustwidth}{0pt}{-9pt}
        \includegraphics[width=\dimexpr\linewidth+9pt\relax]{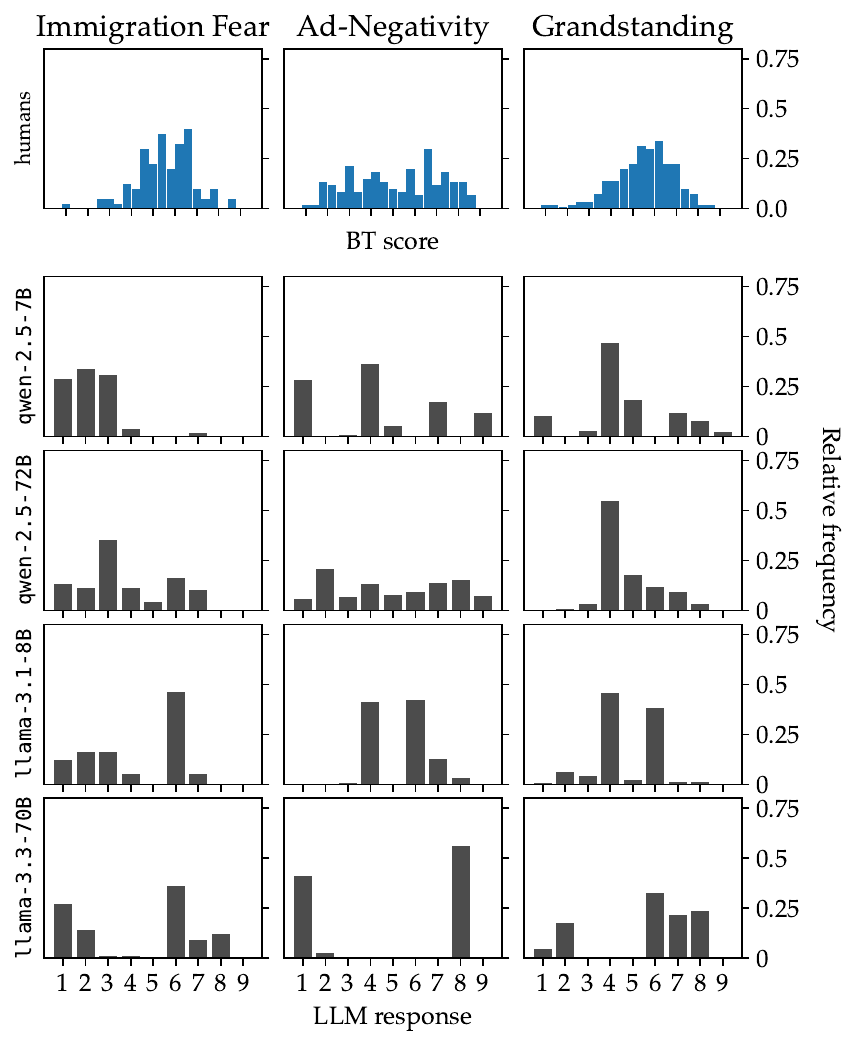}%
    \end{adjustwidth}
    \caption{%
        Distributions of LLM scores for scalar constructs do not align with the reference distribution, nor do they correspond between models.
        \emph{Top:} Distribution of text items' scores on latent dimension for three different tasks estimated by fitting a Bradley-Terry (BT) model to human-annotated pairwise comparisons between text items.
        \emph{Bottom:} Distribution of the scores different LLMs' assign to the same text items if prompted to score them on a 1--9 scale.
}\label{fig:llm_response_distribution_zeroshot}
\end{figure}

\noindent While many constructs in the social sciences are treated as categorical, such as the \textsc{topic} of a speech, %
others are more appropriately considered as a continuum, like the \textsc{emotional intensity} of that speech \citep[e.g.][]{gennaro_emotion_2022, Bagdon2024YouAA}.
Valid scalar measurement of such constructs enables a wide range of substantive applications in social science research, such as modeling legislator behavior \cite{poole2001d} or analyzing polarization in immigration debates \cite{card2022computational}. %
Assigning scalar values to texts is therefore a fundamental task in computational text analysis \cite{Grimmer_Stewart_2013}.

The standard array of NLP methods have been brought to bear on this measurement problem: supervised methods using bag-of-words representations \cite{laver2003extracting,Gentzkow2019MeasuringSpeech}; unsupervised models that assume a latent variable corresponding to the construct \cite{Monroe2004TalksIdeal-points,slapin2008scaling,vafa_text-based_2020,pmlr-v162-hofmann22a,stoehr_sentiment_2023}; and large language models \citep[LLMs; e.g.][]{rottger_political_2024, le_mens_positioning_2025, kim_linear_2025}.

LLMs in particular are an attractive solution for assigning scalar measurements to texts, because in-context learning \cite{brown_language_2020} requires little or no task-specific training data.
However, the space of possible approaches to scoring texts with LLMs is large, and naive prompting can lead to unreliable results \citep{wang_my_2024, rottger_political_2024}.
Take the zero-shot setting, where an LLM is instructed to score texts on an ordinal scale (e.g., 1--9).
Models tend to produce ``heaped'' distributions for this prediction task, wherein probability mass is concentrated only on a few numeric tokens (\cref{fig:llm_response_distribution_zeroshot}). 
This behavior is likely due to systematic biases favoring certain tokens induced during pre- or post-training  \cite{zhao_calibrate_2021, razeghi-etal-2022-impact}.\footnote{Round numbers are far more common: in the Dolma \cite{soldaini-etal-2024-dolma} pretraining set, the n-gram \texttt{25 percent} appears roughly 5M times, compared to about 1M for \texttt{24} or \texttt{26 percent}, per the WIMBD tool from \citet{elazar2024Big}}%
This behavior may mislead researchers studying the absolute level of, or distance between, observations of that construct, pointing to the need to explore alternative approaches.

One such alternative is the \emph{pairwise ranking} of items, where the abstract construct is operationalized as a per-item latent variable that generates observed ranks.
As in the case of human annotation, an LLM compares pairs of texts in terms of their intensity on an underlying scale \citep{patrickcgcot, stoehr-etal-2024-unsupervised}.
The latent per-item scores are then estimated with probabilistic models of ranked pairs, like that from \citet{bradley-terry-1952}.

With human coders, pairwise comparisons produce text rankings that are more robust than annotators' direct ratings of individual text items (\citealt{kendall1948rank,kingsley_preference_2010,de_bruyne_annotating_2021,narimanzadeh2023crowdsourcing}; cf. \citealt{wood_comparison_2018}).
Accordingly, pairwise comparison has been applied to measure various constructs in social science research %
\citep{benoit_measuring_2019, carlson_pairwise_2017} or to validate such measures \citep{gennaro_emotion_2022, hargrave_no_2022}. %
In NLP, pairwise (or listwise) comparisons are common in human annotation \cite{lopez-2012-putting,sakaguchi-etal-2014-efficient,sakaguchi_efficient_2018,simpson-gurevych-2018-finding,chen2021goldilocks,de_bruyne_annotating_2021,narimanzadeh2023crowdsourcing,qin_large_2023,stoehr-etal-2024-unsupervised}, automated system evaluation \cite{liusie-etal-2024-llm,zheng_judging_2023}, and preference modeling \citep{ziegler_fine-tuning_2019,ouyang_training_2022}.

Despite the appeal of pairwise comparisons for measuring social science constructs, we lack comparative evidence of its utility in automated scalar measurement with LLMs \citep[cf.][]{Bagdon2024YouAA}.
We therefore present a comprehensive evaluation of LLM prompting and finetuning methods for text scoring.\footnote{%
    We release all code and data at \url{https://github.com/haukelicht/scalar_measurement}.
}
Using three human-labeled datasets from two political science studies covering different target constructs \citep{carlson_pairwise_2017, park_when_2021}, we consider various prompting methods and calibration techniques---combinations of direct scoring and pairwise comparisons, drawing from the literature on LLM evaluators \cite{wang2025improvingllmasajudgeinferencejudgment}.
In addition to prompting, we adapt reward modeling methods \cite{ziegler_fine-tuning_2019,ouyang_training_2022} to finetune models on pairwise data, comparing them to standard regression finetuning.

We find that the benefit of pairwise comparisons depends on whether the models are prompted or finetuned.
For in-context learning, direct pointwise scoring of text items can be just as (or more) effective than pairwise comparison (\cref{tab:main_results_promnpting}), but only after computing a probability-weighted average over the ordinal tokens \citep[see][]{wang2025improvingllmasajudgeinferencejudgment}.
However, fine-tuning a reward model with as few as 1,000 labeled pairs can produce scoring models that outperform a prompted model (\cref{tab:finetuning_results}), even when the prompted model has two orders of magnitude more parameters (finetuned \emph{regression} models, on the other hand, require more data).

Summarizing our contributions, we:
\begin{compactitem}
\item Analyze issues with direct pointwise scoring and LLMs' ``heaped'' responses (\S\ref{sec:pointwise}).
\item Compile a text scoring benchmark of datasets from the social science literature (\S\ref{sec:data}).
\item Evaluate a suite of LLMs in zero-, few-shot, and fine-tuning settings, comparing pairwise and pointwise scoring (\S\ref{sec:results}).
\item Provide actionable recommendations for practitioners (\S\ref{sec:results}).
\end{compactitem}

\section{Scoring Text Items}

We compare common approaches to scoring text items with LLMs. %
First, we cover \emph{pointwise scoring} via in-context learning (ICL), where the model scores \emph{individual texts} in isolation, and discuss shortcomings of this setup.
We then turn to \emph{pairwise comparisons} with ICL, where the model compares \emph{pairs of texts}.
Last, we discuss finetuning procedures to apply when training data is available.

\subsection{Pointwise Prompting with LLMs}\label{sec:pointwise}

\begin{figure}[t]
    \centering
    \includegraphics[width=\columnwidth]{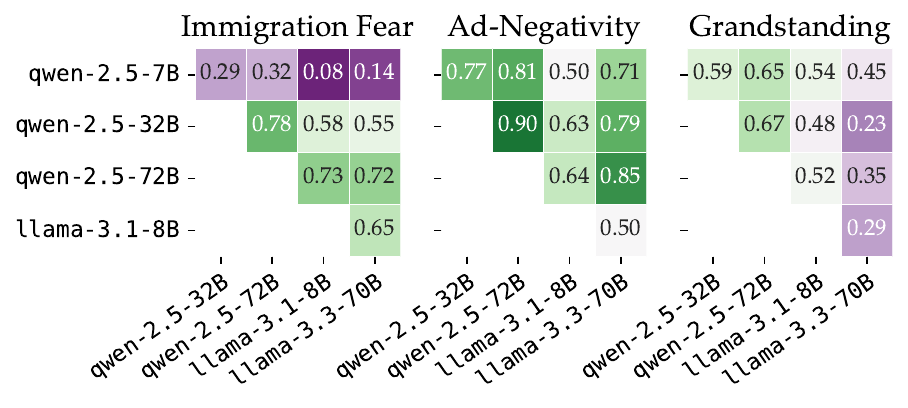}%
    \caption{%
        Inter-model agreement (Krippendorff's $\alpha$) by dataset for zero-shot pointwise prompting. 
        Even within model families, agreement can be relatively low.
    }
    \label{fig:llm_response_agreement_zeroshot}
\end{figure}

Generative LLMs define conditional probability distributions $P(y \mid \mathbf{c})$, where $\mathbf{c} \in \Sigma^{*}$ is a prompt, and $y \in \Sigma$ is the next token from the vocabulary $\Sigma$ to be generated.
The most straightforward way to score a text item $\mathbf{x}_i$, therefore, is to prompt an LLM to output a value on a fixed scale, such as \mbox{1--9}~\cite{tian-etal-2023-just, 
ohagan2024measurementagellmsapplication,ziems-etal-2024-large,le_mens_positioning_2025,Bagdon2024YouAA}.
The researcher defines the scale in terms of discrete answer token candidates $\mathcal{S}\subset\Sigma$ (e.g., $\mathcal{S} = \{\texttt{1}, \ldots, \texttt{9}\}$) and instructs the LLM to assign a text item one of the available scale point values, per a prompt $\mathbf{c}_i$.
This \emph{pointwise} prompting strategy can be further refined by including coding instructions in the prompt~\cite{ruckdeschel-2025-just}, or by averaging scores over multiple sentences within a document~\cite{le_mens_positioning_2025}.

However,~\citet{wang2025improvingllmasajudgeinferencejudgment} show that instead of relying on an LLM's most-probable response $\argmax_y P(y\mid \mathbf{c}_i)$), the weighted average of a model's token probabilities produces more accurate scores in an LLM-as-judge setting (following up on findings from \citealt{liu-etal-2023-g,yasunaga2024almaalignmentminimalannotation,lee2024RLAIF}).
The score for a text item $\mathbf{x}_i$ is then:
\begin{align}
    s_i & = \frac{1}{p_s}\sum_{s \in \mathcal{S}} P(y=s \mid \mathbf{c}_i) \cdot n(s),\label{eq:token-weighted-mean}
\end{align}
with $p_s=P(y \in \mathcal{S} \mid \mathbf{c}_i)$, the total probability mass assigned to tokens in the scale $\mathcal{S}$, and $n:~\mathcal{S}\rightarrow~\mathbb{Z}$, a function mapping tokens in the scale to their corresponding integer representations.

\paragraph{Pitfalls of pointwise prompting.}
Prompting LLMs to directly score individual text items has several limitations \citep{ohagan_measurement_2023}.
First, the scores that generative LLMs assign to individual text items tend to be poorly calibrated: common token bias \citep{zhang2024forcingdiffusedistributionslanguage} and prompt phrasing \cite{sclar2024prompt} can dramatically affect models' responses.
See \Cref{fig:llm_response_distribution_zeroshot}: 
for the three datasets we study (\cref{sec:data}), the zero-shot pointwise scoring outputs of different LLMs produce distributions over items that do not agree with scores inferred from ground-truth human annotations using Bradley-Terry (see \cref{sec:pairwise}).

Consider the results for the \textsc{Immigration Fear} data, which focuses on survey respondents' anxieties about immigration in the U.S. \citep[see \cref{sec:data}]{carlson_pairwise_2017}.
The reference distribution is centered around the midpoint of the inferred scale, bimodal, and symmetric.
None of the distributions of LLM scores align with this reference.
The responses of Qwen 2.5 models \cite{qwen2025qwen25technicalreport}, for example, tend to be right-skewed, especially for the smaller 7B variant.
And while both Llama 3 models' \cite{dubey2024llama3herdmodels} responses are bimodal, they do not match the symmetry in the reference distribution.

\Cref{fig:llm_response_distribution_zeroshot} also illustrates a second pathology. When LLMs are asked to score texts, their responses can exhibit a phenomenon known as \emph{heaping}, where model outputs are concentrated on particular values, rather than using the full extent of the scale \citep[see][]{roberts_measures_2001}.\footnote{
    Heaping is especially likely when using fine-grained scales, such as 0--100, because models often choose scores divisible by 10 or 5 \citep[e.g.,][]{le_mens_positioning_2025}.
    For LLMs, heaping can be caused by a lack of explicit diversity incentives during training or bias in the training data~\cite{zhang2024forcingdiffusedistributionslanguage}.
}
For example, \llamabig\ outputs scores $y = \texttt{1}$ and $y = \texttt{8}$ for most of the text items in the \textsc{Ad-Negativity} data.

\begin{figure}[t]
    \centering
    \includegraphics[width=\columnwidth]{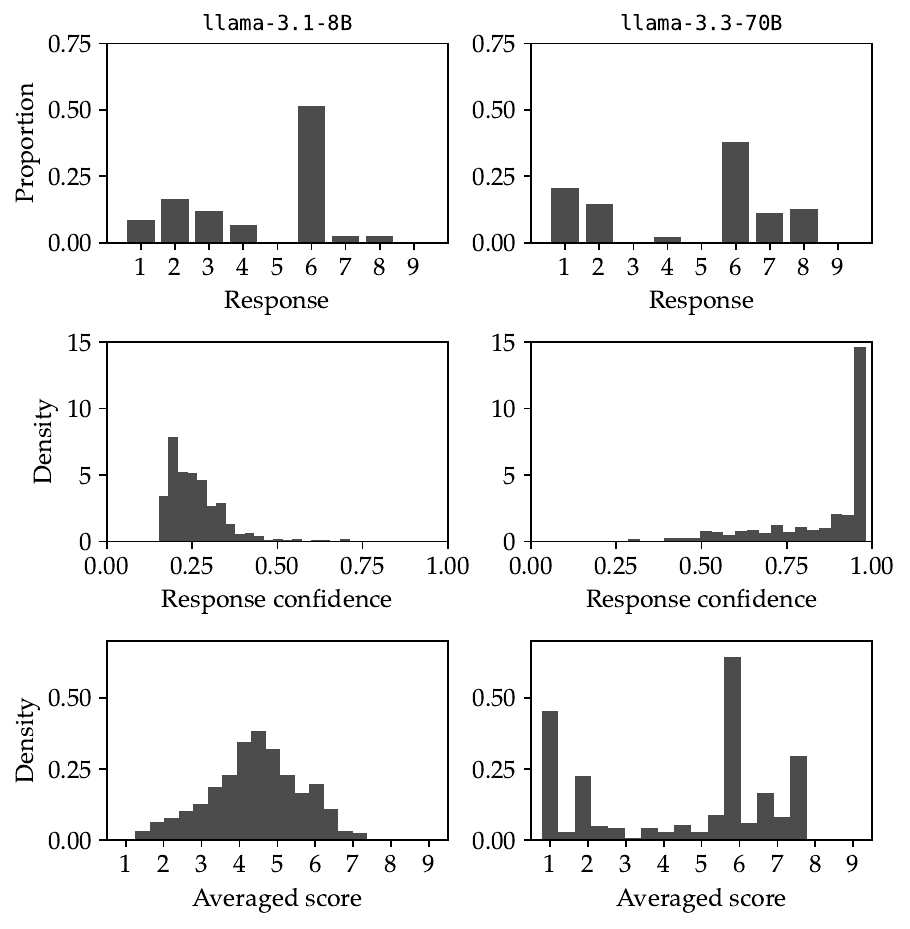}%
    \caption{%
       Distribution of \llamasmall\ and \llamabig's modal responses (top row), confidences in modal responses (mid row), and probability-weighted average responses bottom row) from zero-shot pointwise prompting for texts in the \textsc{Immigration Fear} data (see \S\ref{sec:data}).
       While both LLMs modal response distributions exhibit similar levels of concentration on $y=6$, the larger model variant (right) tends to be moch more confident its response, which reduces the smoothing effect probability-weighted averaging has on its response distribution.
   }
\label{fig:prompting_response_confidence_averaging_effect}
\end{figure}

Of course, not all these response distributions can be true at the same time, undermining the reliability of LLMs' zero-shot pointwise scoring responses.
Accordingly, models' responses often agree at best moderately, even within a model family (see \cref{fig:llm_response_agreement_zeroshot}).

Calibration could potentially be improved by taking the probability-weighted average over numerical tokens in the scale (eq. \ref{eq:token-weighted-mean}), as it could smooth the distributions, rendering them more like the reference.
However, LLMs' confidence over tokens is often poorly calibrated and heavily concentrated on the modal response \citep{tian-etal-2023-just,xie-etal-2024-calibrating}, which can distort the resulting weighted average (\cref{fig:prompting_response_confidence_averaging_effect}). %
In addition, confidences are strongly affected by the model size and prompting method (\cref{fig:prompting_response_confidence_distribution} in the Appendix).

Although prior work suggests strong correlations between pointwise LLM scores and human ground truth at the document level for certain constructs \cite{le_mens_positioning_2025}, the above results indicate that LLMs are not a panacea. %
In the remainder of this work, we systematically assess the reliability of scoring text via pointwise prompting, pairwise prompting, and finetuning.

\subsection{Pairwise Comparisons}\label{sec:pairwise}

An alternative to pointwise scoring is collecting pairwise preferences. 
Given a pair of text items, a human annotator or LLM selects the item that better exemplifies the target construct or that is more extreme on the underlying dimension.  
Given multiple such comparisons, a probabilistic pairwise ranking model such as Bradley-Terry~\citep[][henceforth referred to as BT]{bradley-terry-1952} can be used to estimate items' location on the latent construct.\footnote{A primer on Bradley-Terry is in \Cref{apx:bradley-terry}.}
The BT model takes the pairwise outcomes as input and estimates the latent ``strength'' of each text item via maximum likelihood, modeling the win probability with a logistic link function:\footnote{There are many probabilistic rank models \citep{mallows_non-null_1957, luce1959individual, elo_proposed_1967, plackett1975,herbrich2006trueskill,lu_learning_2011,carlson_pairwise_2017}; we focus on Bradley-Terry due to its simplicity and ubiquity.
}
\begin{align}
    P(i > j) = \frac{e^{z_i}}{e^{z_i}+e^{z_j}}.
\end{align}
for texts $\mathbf{x}_i, \mathbf{x}_j$ and their corresponding latent scores $z_i, z_j$.%

For different social science constructs measured with human annotations, scores inferred from annotators' pairwise comparisons measure the target dimensions more reliably than ratings obtained from expert annotations through direct, pointwise annotation \citep{carlson_pairwise_2017}. 
More recently, pairwise judgments obtained from LLMs have also been leveraged to estimate constructs in social science applications~\cite{sarkar-etal-2025-pairscale, patrickcgcot, Bagdon2024YouAA}.\footnote{There is also a computational downside: direct scoring is $\mathcal{O}(n)$, whereas pairwise is $\mathcal{O}(n^2)$, although subsampling can improve this to $\mathcal{O}(kn)$. }
However, a gap remains in the literature: studies using pairwise comparisons often lack adequate comparisons to pointwise scoring baselines~\cite{sarkar-etal-2025-pairscale}, and vice versa~\cite{le_mens_positioning_2025,ohagan2024measurementagellmsapplication}.\footnote{The study by \citet{Bagdon2024YouAA} is an exception but they focus on a single construct: emotion intensity.}

\begin{figure}[t]
    \centering
    \includegraphics[width=.8\columnwidth]{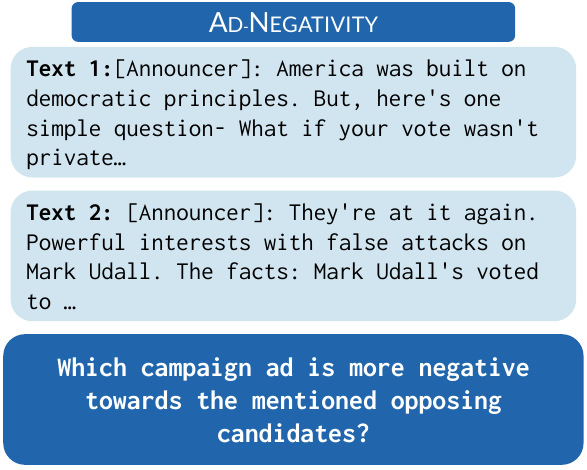}%
    \caption{Pairwise comparison places two text items relative to one another regarding a given construct.}
    \label{fig:data}
\end{figure}

\subsection{Finetuning}

If labeled data are available, another option is \emph{finetuning} LLMs \citep{howard-ruder-2018-universal}.
When the labeled data are annotated pairs (as is the case here, see \S\ref{sec:data}), we adapt \emph{reward modeling} objectives, used to align language models to human preferences with pairwise data \citep[][]{christiano_deep_2017,ziegler_fine-tuning_2019,ouyang_training_2022}.
Per \citeauthor{ouyang_training_2022}, the loss is the negative-log likelihood of a pairwise comparison under Bradley-Terry,
\begin{align}
    \label{eq:reward_modeling}%
    \ell_\theta(\mathbf{x}_{h}, \mathbf{x}_{l}) = - \log \left(\sigma \left(r_\theta(\mathbf{x}_h)  -  r_\theta(\mathbf{x}_l)\right)\right),
\end{align}
with $\mathbf{x}_{h}$ being the preferred item in the pair (over $\mathbf{x}_l$), and $r_\theta(\cdot)$ being an LLM with a regression head.
The score for an item $\mathbf{x}_i$ is then $r_\theta(i)$ (meaning no pairwise comparisons are required at inference time). The model is trained via gradient descent.
An alternative is regression on available scores directly, for example with a mean squared error loss.

\section{Experimental Setup}

Given the established benefits of using pairwise annotations in social science data, our datasets all consist of pairwise comparison data from the social science literature.
First, we outline each dataset (statistics in \cref{tab:dataset_descriptives}), then discuss methods, LLM variants, evaluation strategy,  and metrics.

\subsection{Datasets}\label{sec:data}

\paragraph{\ImmFear}  \citet{carlson_pairwise_2017} rely on trained crowdworkers to measure \emph{the level of fear, anxiety, or worry toward immigration or immigrants in the U.S.} expressed in responses to an open-ended survey question.
The construct targeted in this dataset relates to other research on the use of (discrete) emotions in political communication \citep{widmann_creating_2022,gennaro_emotion_2022}.

\paragraph{\AdNeg}
In the same work, \citeauthor{carlson_pairwise_2017} also crowd-source pairwise comparisons to measure the \emph{level of negativity of political campaign advertisements}, a construct related to negative campaigning and other attack behaviors \citep{walter_explaining_2015,licht_measuring_2025}.
The ads in their data were aired before the 2008 U.S.~Senate elections and obtained from the Wisconsin Advertising Project (WiscAds) database.\footnote{\url{elections.wisc.edu/wisconsin-advertising-project/}}
The annotations come from trained online workers, who indicate which ad in a pair is ``most negative towards'' or ``least positive about'' the ``candidate(s) mentioned'' \citep[p.~828]{carlson_pairwise_2017}.\footnote{%
    \citet{ornstein_how_2025} use experts' ad-level (pointwise) ratings to evaluate LLMs' pointwise scoring in this data.
}

\paragraph{\Grand}
Last, we use a dataset compiled by \citet{park_when_2021}, who measures speakers' \emph{grandstanding} in House committee hearings in the U.S. Congress. 
Park defines grandstanding as opinionized speech behavior that ``sends political messages by taking positions on policy issues or framing the image of a party or the administration'' and contrasts it with information-seeking, fact-oriented speech behavior.
The texts are short statements sampled from roughly 12,000 speeches held by House committee members during public hearings in the 105th to 114th Congresses.
Trained online workers indicate which statement in a pair would be better described as opinionized or grandstanding as opposed to fact-based or information seeking.

\begin{table}
\resizebox{\textwidth}{!}{%
\begin{tabular}{lrrrrr}
\toprule
 & & &
 \multicolumn{3}{c}{Node degree statistics}
 \\  \cmidrule(lr){4-6}
 & Items & Pairs & Co. & Dens. & $\mu\ (\sigma)$ \\
\midrule
\textsc{Immigration Fear} & 334 & 6,489 & 33 & 0.11 & 37.6 (1.5)  \\
\AdNeg & 935 & 9,489 & 18 & 0.02 & 20.2 (0.9)  \\
\Grand & 3,499 & 38,348 & 17 & 0.01 & 21.8 (7.6)  \\
\bottomrule
\end{tabular}
}
\caption{
  Datasets consist of annotated pairs of items, forming a graph. 
  Connectivity (Co.) is the number vertices (\emph{items}) needed to disconnect the graph; 
  density (Dens.) is the ratio of observed edges (\emph{pairs}) to possible edges.
  $\mu$ ($\sigma$) are the mean (std.\,dev.) degree per node.
}\label{tab:dataset_descriptives}
\end{table}

\subsection{Methods and Models}\label{sec:models}
\paragraph{Prompting.} Prompted models infer either pointwise scores per item or pairwise ranks per pair.
For pointwise prompting, we instruct the model to provide an integer on a 1--9 scale per construct.\footnote{We use a range between 1 and 9 because (some) models tokenize two-digit numbers into two tokens.}
To produce a final score, we apply the probability weighting from \Cref{eq:token-weighted-mean}.

In the pairwise case, the model selects the item that is more extreme for the construct.
We borrow from~\citet{wang2025improvingllmasajudgeinferencejudgment} and prompt the model with different orderings to avoid positional biases \citep{zhao_calibrate_2021, han_prototypical_2023, wang_large_2023}.
We record the distribution over the choice tokens and take the average (see \Cref{apx:pairwise_debiasing}).
In both settings, we run zero-shot and random 5-shot exemplars.\footnote{For pointwise scores, using anchors for the different scale points did not consistently improve metrics over 5-shot.} %
The instructions are derived from the original publications' codebooks (prompts in \Cref{apx:prompts}).

We use instruction-tuned open-weight LLMs in different sizes:  Llama 3.1 8B, Llama 3.3 70B  \citep{dubey2024llama3herdmodels} as well as the 7B, 32B, and 72B variants of Qwen 2.5 \citep{qwen2025qwen25technicalreport}.
All models were run with 4-bit quantization.\footnote{
    Using the \texttt{transformers} \citep{wolf_transformers_2020} and \texttt{bitsandbytes} \citep[see][]{dettmers_qlora_2023}.
    Our results are robust to using no quantization (\cref{tab:no_quant_prompting_results}).
    GPU specifications are reported in \Cref{tab:gpu_specs}.
}

\paragraph{Finetuning.} 
Using the reward modeling objective in~\cref{eq:reward_modeling}, we finetune \deberta\ \citep{he2021DeBERTav3}, \modbert\  \citep{warner_smarter_2024}, and \llamai\ models.
The first two models are encoder-only models commonly used to fine-tune classifiers in political text analysis \citep[cf.][]{timoneda_bert_2024}; Llama is a decoder-only generative language model with double the parameters of \deberta.
To finetune the 8B Llama model, we use QLoRA with 4-bit quantization \citep{dettmers_qlora_2023}.
We sweep over the learning rate for each model to maximize pairwise accuracy on a validation set (see \Cref{apx:finetuning_hparams}).

Further, to compare pairwise to pointwise finetuning, we finetune \deberta\ with a regression head. The labels are scores estimated from BT models fit to the human pairwise annotations in the training set.

\begin{table*}[ht]
\caption{%
    Comparison of pointwise scoring and pairwise comparison LLM prompting methods for scalar construct measurement.
    Item-level scores in pairwise comparison setup are inferred with Bradley-Terry (BT); pointwise scores use token probability-weighted averaging.
    Pointwise scoring tend to work better and is computationally cheaper.
    \emph{Notes:}
    Results reported for 0-shot prompting (no exemplars) and 5-shot prompting (using five randomly sampled exemplars per dataset).
    Accuracy (Acc) measured relative to human ground-truth pairwise comparisons; Spearman's rank correlation ($\rho$) and root mean squared error (RMSE) are relative to the ground-truth BT scores inferred from those comparisons.
    Values report averages $\pm$ one standard deviation computed based on 25 bootstrapped estimates in test split.
    Values in bold mark the best result for a dataset and metric and values underlined flag results within one standard error of the best result.
}
\label{tab:main_results_promnpting}
\centering
\small
\resizebox{\textwidth}{!}{
\bgroup
\setlength\tabcolsep{3pt}
\begin{tabular}{rcccccccccc}
\toprule
 & & \multicolumn{3}{c}{\textsc{ Immigration Fear }} & \multicolumn{3}{c}{\textsc{ Ad-Negativity }} & \multicolumn{3}{c}{\textsc{ Grandstanding }} \\\cmidrule(lr){3-5}\cmidrule(lr){6-8}\cmidrule(lr){9-11}
 & Shots & Acc & $\rho$ & RMSE & Acc & $\rho$ & RMSE & Acc & $\rho$ & RMSE \\
\midrule
\addlinespace
\multicolumn{11}{l}{\bfseries\texttt{Qwen-2.5-7B}} \\\addlinespace[1pt]
~~pairwise & 0-shot & 0.65±0.01 & 0.53±0.07 & 0.18±0.01 & 0.76±0.01 & 0.83±0.02 & 0.20±0.01 & 0.62±0.01 & 0.45±0.04 & 0.21±0.01 \\
 & 5-shot & 0.73±0.01 & 0.81±0.03 & \textbf{0.15±0.01} & 0.77±0.01 & 0.84±0.02 & 0.18±0.01 & 0.61±0.01 & 0.40±0.05 & 0.23±0.01 \\
\addlinespace
~~pointwise & 0-shot & 0.70±0.01 & 0.63±0.06 & 0.40±0.01 & 0.78±0.01 & 0.85±0.02 & 0.19±0.01 & 0.65±0.01 & 0.58±0.05 & 0.22±0.01 \\
 & 5-shot & 0.75±0.01 & 0.81±0.04 & 0.26±0.01 & 0.79±0.01 & 0.87±0.02 & 0.20±0.01 & 0.65±0.01 & 0.57±0.05 & 0.24±0.01 \\
\addlinespace
\multicolumn{11}{l}{\bfseries\texttt{Qwen-2.5-32B}} \\\addlinespace[1pt]
~~pairwise & 0-shot & 0.71±0.01 & 0.68±0.04 & 0.26±0.01 & 0.80±0.01 & 0.85±0.02 & 0.18±0.01 & 0.66±0.01 & 0.51±0.04 & 0.20±0.01 \\
 & 5-shot & \textbf{0.77±0.01} & \underline{0.85±0.03} & 0.16±0.01 & \underline{0.80±0.01} & 0.85±0.02 & 0.18±0.01 & \underline{0.68±0.01} & 0.59±0.03 & 0.19±0.01 \\
\addlinespace 
~~pointwise & 0-shot & 0.72±0.01 & 0.71±0.04 & 0.26±0.01 & 0.79±0.01 & 0.87±0.02 & 0.14±0.01 & 0.65±0.01 & 0.58±0.04 & 0.31±0.01 \\
 & 5-shot & \underline{0.77±0.01} & \underline{0.85±0.03} & 0.22±0.01 & \underline{0.80±0.01} & 0.90±0.02 & 0.16±0.01 & 0.66±0.01 & 0.61±0.05 & 0.30±0.01 \\
\addlinespace
\multicolumn{11}{l}{\bfseries\texttt{Qwen-2.5-72B}} \\\addlinespace[1pt]
~~pairwise & 0-shot & 0.73±0.01 & 0.72±0.05 & 0.19±0.01 & \underline{0.80±0.01} & 0.86±0.02 & 0.18±0.01 & 0.66±0.01 & 0.48±0.04 & 0.20±0.01 \\
 & 5-shot & \textbf{0.77±0.01} & 0.84±0.03 & 0.16±0.01 & \underline{0.81±0.01} & 0.87±0.02 & 0.19±0.01 & 0.67±0.01 & 0.56±0.03 & 0.19±0.01 \\
\addlinespace
~~pointwise & 0-shot & 0.76±0.01 & 0.81±0.03 & 0.23±0.01 & \underline{0.81±0.01} & 0.91±0.01 & \textbf{0.13±0.01} & 0.65±0.01 & 0.60±0.04 & 0.21±0.01 \\
 & 5-shot & \underline{0.77±0.01} & \textbf{0.87±0.03} & 0.20±0.01 & \textbf{0.81±0.01} & \textbf{0.92±0.01} & 0.17±0.01 & 0.67±0.01 & \textbf{0.66±0.04} & 0.29±0.01 \\
\addlinespace
\multicolumn{11}{l}{\bfseries\texttt{Llama-3.1-8B}} \\\addlinespace[1pt]
~~pairwise & 0-shot & 0.58±0.01 & 0.48±0.09 & 0.32±0.01 & 0.59±0.01 & 0.46±0.08 & 0.26±0.02 & 0.55±0.01 & 0.21±0.07 & 0.25±0.01 \\
 & 5-shot & 0.73±0.01 & 0.75±0.05 & 0.21±0.01 & 0.74±0.01 & 0.74±0.04 & 0.21±0.01 & 0.59±0.01 & 0.36±0.06 & 0.23±0.01 \\
\addlinespace
~~pointwise & 0-shot & 0.71±0.01 & 0.68±0.06 & 0.18±0.01 & 0.77±0.01 & 0.82±0.02 & 0.24±0.01 & 0.64±0.01 & 0.57±0.04 & \textbf{0.16±0.01} \\
 & 5-shot & 0.74±0.01 & 0.78±0.04 & 0.27±0.01 & \underline{0.81±0.01} & 0.90±0.01 & 0.20±0.01 & 0.66±0.01 & 0.61±0.05 & 0.39±0.01 \\
\addlinespace
\multicolumn{11}{l}{\bfseries\texttt{Llama-3.3-70B}} \\\addlinespace[1pt]
~~pairwise & 0-shot & 0.73±0.01 & 0.74±0.05 & 0.18±0.01 & 0.78±0.01 & 0.83±0.03 & 0.19±0.01 & 0.66±0.01 & 0.54±0.03 & 0.19±0.01 \\
 & 5-shot & 0.76±0.01 & \underline{0.85±0.03} & \underline{0.15±0.01} & 0.79±0.01 & 0.84±0.02 & 0.19±0.01 & 0.67±0.01 & 0.54±0.03 & 0.20±0.01 \\
\addlinespace
~~pointwise & 0-shot & 0.73±0.01 & 0.75±0.04 & 0.30±0.01 & 0.80±0.01 & 0.89±0.01 & 0.29±0.01 & 0.66±0.01 & 0.55±0.04 & 0.26±0.01 \\
 & 5-shot & 0.76±0.01 & 0.84±0.03 & 0.28±0.01 & \underline{0.80±0.01} & 0.90±0.02 & 0.25±0.01 & \textbf{0.69±0.01} & \underline{0.65±0.04} & 0.30±0.01 \\
\bottomrule
\end{tabular}
\egroup
}
\end{table*}

\subsection{Evaluation Strategy}\label{sec:evaluation_strategy}

Each dataset contains a unique set of text \emph{items} $\mathbf{x}_i$; labels between pairs indicate which item is more extreme on the target scalar construct (e.g., which of two campaign ad transcripts is more negative about an opponent).
Therefore, the pairwise labeled data in each dataset form a directed connected graph $G = (V, E)$ with vertices representing text items and edges representing comparison relations.\footnote{
    This structure is unlike standard LLM paired preference datasets, which are disconnected. There, comparisons between generated texts are conditioned on some shared context, like an instruction, making them incomparable across contexts.
}

The graphs are connected, so constructing a train--test split and an adequate evaluation strategy is a somewhat delicate operation.
We proceed as follows.
First sample $n_\text{test}=100$ high-degree %
held-out evaluation vertices $V_\text{test} \in V$, with train vertices $V_\text{train} \coloneqq V\setminus V_\text{test}$.
The induced subgraph $G_\text{train} \coloneqq G\left[V_\text{train}\right]$ comprises the edges and items used for training the finetuned models.
$G_\text{eval} \coloneqq G\left[E \setminus E_\text{train}\right]$ is then the graph of all edges that contain at least one test vertex $V_\text{test}$ (i.e., the graph also contains items from $V_\text{train}$).

\paragraph{Evaluation metrics.}
Our primary focus is on LLMs' \emph{scoring performance}.
To estimate items' ground-truth reference scores, we fit a Bradley-Terry (BT) model\footnote{Using the \texttt{choix} implementation \citep{maystre_choix_2022}} \emph{to the entire graph} $G$, resulting in item-level scores $r_{\text{BT}}(\mathbf{x}_i)$ for all $\mathbf{x}_i \in V$.
We then evaluate how well an LLM can predict these item-level scores.
Specifically, we measure scoring performance with \textbf{Spearman rank correlation} $\bm{\rho}$ and the root mean squared error (\textbf{RMSE}) in the subset of items in $V_\text{test}$.\footnote{%
This avoids leakage when evaluating finetuned models and makes evaluations of prompted and finetuned models comparable.
}
$\rho$ measures how well an LLM ranks text items relative to the reference scores.
RMSE measures the average magnitude of the errors between text items' predicted and reference scores. %
These metrics are complementary: $\rho$ focuses on ranking consistency, while RMSE evaluates the precision of the numerical predictions.

A secondary focus is on \emph{pair classification performance}, i.e., whether models can predict the pairwise labels $\mathbb{I}[\mathbf{x}_i>\mathbf{x}_j],\ (\mathbf{x}_i,\mathbf{x}_j)\in E$.
We measure classification performance with \textbf{accuracy}, which we compute against \emph{all} edges in $G_\text{eval}$---that is, any edge where a test vertex appears.\footnote{This choice biases the finetuned accuracies upward because they have seen the train vertices, but there are relatively few edges with both $\mathbf{u}, \mathbf{v} \in V_\text{test}$.}

\paragraph{Influence of training data size.}
For the finetuning experiments, we also evaluate the effect of increasing the number of edges, while controlling for differences in dataset structure (\cref{tab:dataset_descriptives}).
Our algorithm iteratively adds edges to each of the three graphs such that the average degree and clustering coefficient remain the same for each $n$, up until about 2000 edges, where maintaining graph similarity is no longer feasible given the differing structural characteristics of our datasets.

\section{Results}\label{sec:results}

First, we report results from prompting and then fine-tuning. 
Overall, pairwise comparison does not improve prompting results (see \cref{tab:main_results_promnpting}), but it does help in finetuning (\cref{fig:finetuning_train_size_vs_regression}).

\subsection{Prompting}
\Cref{tab:main_results_promnpting} compares pairwise and pointwise prompting approaches for text scoring.
We report results for both zero- and 5-shot prompting %
 with probability-weighted averaging.\footnote{
     Using probability-weighted averages instead of models' modal response tends to improve the accuracy and scoring performance
     for all models and few-shot settings in the \Grand\ data and the \AdNeg\ data (but one)
     and 
     for most models and most few-shot settings in the \ImmFear\ data for accuracy and RMSE.
}

\begin{table*}[ht]
\caption{Reward model finetuning results by model, number of training examples, and dataset. Finetuning tends to outperform prompting after roughly 2,000 examples. \emph{Notes:} Metrics reported are the pair classification accuracy (Acc), as well as the Spearman's rank correlation ($\rho$) and the root mean squared error (RMSE) against the ground-truth BT scores. Values are averages $\pm$ one standard deviation computed by summarizing results across five folds.}
\label{tab:finetuning_results}
\small
\resizebox{\textwidth}{!}{
\bgroup
\setlength\tabcolsep{3pt}
\begin{tabular}{rrccccccccc}
\toprule
\multicolumn{2}{c}{} & \multicolumn{3}{c}{\textsc{ Immigration Fear } (\emph{all} = 2729)} & \multicolumn{3}{c}{\textsc{ Ad-Negativity } (\emph{all} = 6760)} & \multicolumn{3}{c}{\textsc{ Grandstanding } (\emph{all} = 32308)} \\\cmidrule(lr){3-5}\cmidrule(lr){6-8}\cmidrule(lr){9-11}
 & $N_{\text{train}}$ & Acc & $\rho$ & RMSE & Acc & $\rho$ & RMSE & Acc & $\rho$ & RMSE \\
\midrule
\addlinespace
\multicolumn{11}{l}{\bfseries \texttt{DeBERTa-v3-large}}  \\\addlinespace[3pt]
 & 500 & 0.74±0.02 & 0.79±0.10 & 0.28±0.02 & 0.78±0.01 & 0.86±0.02 & 0.23±0.02 & 0.64±0.01 & 0.67±0.04 & 0.22±0.04 \\
 & 1000 & 0.75±0.01 & 0.83±0.06 & 0.23±0.05 & \underline{0.80±0.01} & 0.88±0.01 & 0.20±0.05 & 0.65±0.01 & 0.70±0.05 & 0.21±0.05 \\
 & 2000 & 0.77±0.01 & \underline{0.89±0.03} & \underline{0.19±0.01} & \underline{0.80±0.01} & 0.89±0.02 & 0.19±0.05 & 0.66±0.01 & 0.73±0.03 & 0.22±0.09 \\
 & \emph{all} & \textbf{0.78±0.00} & \textbf{0.91±0.02} & \underline{0.18±0.04} & \textbf{0.81±0.01} & \textbf{0.91±0.00} & 0.19±0.03 & \textbf{0.68±0.01} & \textbf{0.78±0.03} & \textbf{0.16±0.02} \\
\addlinespace
\multicolumn{11}{l}{\bfseries \texttt{ModernBERT-large} } \\\addlinespace[3pt]
 & 500 & 0.74±0.01 & 0.79±0.04 & \underline{0.20±0.05} & 0.74±0.02 & 0.77±0.04 & 0.20±0.04 & 0.61±0.03 & 0.51±0.07 & 0.22±0.03 \\
 & 1000 & 0.76±0.01 & 0.83±0.02 & \underline{0.20±0.07} & 0.75±0.01 & 0.80±0.03 & 0.19±0.05 & 0.62±0.01 & 0.56±0.03 & 0.19±0.03 \\
 & 2000 & 0.77±0.01 & 0.85±0.03 & \textbf{0.17±0.04} & 0.78±0.02 & 0.83±0.04 & \underline{0.17±0.04} & 0.64±0.01 & 0.62±0.05 & 0.19±0.05 \\
 & \emph{all} & 0.77±0.01 & 0.85±0.01 & \underline{0.18±0.04} & 0.79±0.02 & 0.87±0.03 & \underline{0.17±0.04} & 0.66±0.01 & 0.72±0.03 & 0.21±0.05 \\
\addlinespace
\multicolumn{11}{l}{\bfseries \texttt{ Llama-3.1-8B-Instruct} } \\\addlinespace[3pt]
 & 500 & 0.76±0.01 & 0.86±0.04 & \underline{0.20±0.08} & 0.78±0.01 & 0.85±0.02 & \textbf{0.16±0.02} & 0.64±0.01 & 0.63±0.06 & 0.20±0.03 \\
 & 1000 & 0.77±0.01 & 0.86±0.03 & \underline{0.19±0.04} & \underline{0.80±0.01} & 0.88±0.01 & \underline{0.18±0.06} & 0.63±0.02 & 0.60±0.09 & 0.19±0.03 \\
 & 2000 & 0.76±0.01 & 0.82±0.03 & 0.23±0.02 & 0.79±0.01 & 0.85±0.05 & \textbf{0.16±0.03} & 0.62±0.00 & 0.60±0.04 & 0.19±0.03 \\
 & \emph{all} & 0.76±0.01 & 0.83±0.05 & \textbf{0.17±0.04} & 0.79±0.01 & 0.87±0.02 & \underline{0.17±0.05} & 0.65±0.01 & 0.70±0.05 & 0.21±0.04 \\
\addlinespace\midrule
\multicolumn{11}{l}{ \emph{best pointwise prompting results} (see Table 2)} \\\addlinespace
 &  & 0.77±0.01 & 0.87±0.03 & 0.18±0.01 & 0.81±0.01 & 0.92±0.01 & 0.13±0.01 & 0.69±0.01 & 0.66±0.04 & 0.16±0.01 \\
\bottomrule
\end{tabular}
\egroup
}
\end{table*}

\paragraph{Pointwise outperforms pairwise prompting.}
Across all three datasets, 5-shot pointwise prompting with probability-weighted averaging consistently outperforms or is equally reliable as zero- or 5-shot pairwise prompting (table \ref{tab:main_results_promnpting}).\footnote{
    Using a proprietary model (GPT-4o) yields comparable results (\cref{tab:gpt4o_results}).
}
This denotes a key difference in human and LLM construct measurement.
Humans demonstrate increased accuracy with pairwise comparisons over pointwise scoring due to reduced calibration errors~\cite{carlson_pairwise_2017}.
This advantage does not translate to LLMs in our datasets, however.
Unsurprisingly, it is the larger model variants that achieve these top scores (\qwenbig\  for \ImmFear\ and \AdNeg, and \llamabig\  for \Grand).
In line with previous research~\cite{chamieh-etal-2024-llms}, 5-shot prompting substantially improves the pointwise scoring performance in terms of accuracy and correlation. 
The magnitude of improvement depends on model size (smaller models benefit more) and on the task (few-shot exemplars help more in \Grand\ dataset).\footnote{We experiment with other exemplar selection strategies, but did not see an improvement over choosing five at random.}

\begin{figure}[!t]
    \centering
    \includegraphics[width=\textwidth]{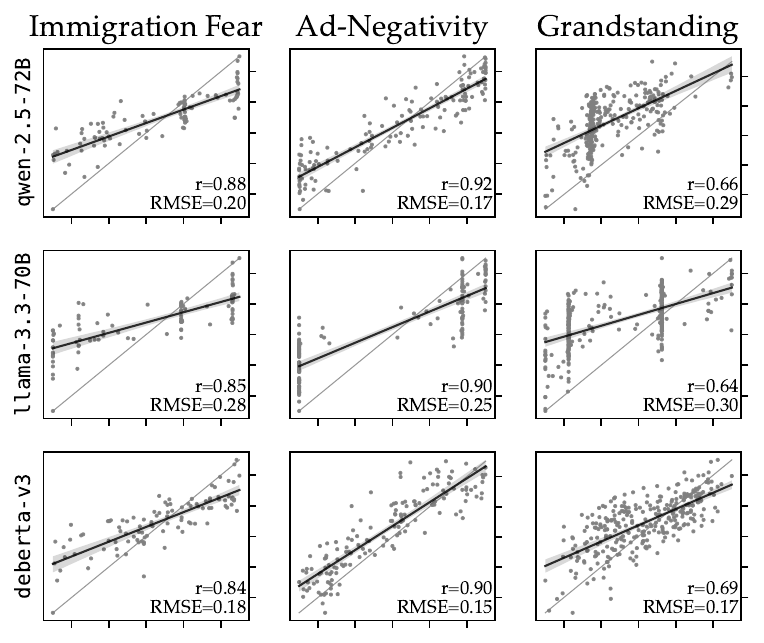}%
    \caption{
        Relation between model responses and ground-truth BT scores.
        \qwenbig\ and \llamabig\ are prompted, showing probability-weighted average pointwise scores (5-shot). 
        \texttt{DeBERTa-v3} (large) has been finetuned on 2,000 training pairs for each datasets.
        These results show RMSE and Spearman's $\rho$ may not always agree.
    }
    \label{fig:scatterplots_combined}
\end{figure}

\paragraph{Correlation values can mask important differences in error magnitude.}
While correlation can be an effective measure of how well a model ranks text items relative to ground truth, looking at correlation alone fails to capture how predicted scores diverge from the true distribution. 
In the \AdNeg\ dataset, both \qwenbig\ and \llamabig\ (with 5-shot prompting) have high correlations.
However, their RMSE scores reveal that \llamabig\ makes larger errors, while achieving a rank correlation similar to that of the other two models (\cref{fig:scatterplots_combined}). 
\llamabig's worse RMSE scores can be attributed to the fact that, due to its extremely high response confidence (\cref{fig:prompting_response_confidence_distribution}),  the heaping in its modal responses (\cref{fig:llm_response_distribution_zeroshot}) is virtually unaffected by token probability weighting (\cref{fig:prompting_response_confidence_averaging_effect}).
Consequently, its scores are bunched around 0.0, 0.9, and 1.0.

This example underscores that, for scoring tasks, considering both RMSE and correlation provides a balanced view of model behavior. 
Specifically, the choice of a model should depend on a researcher's research design.
If their analysis only requires that texts are sorted in the correct order, the strong tendency to produce heaped outputs of \llamabig\ should be less of a concern.
However, if their research design requires measuring the relative distances between observations, RMSE becomes more important, and \qwenbig\ might be a better option.

\subsection{Finetuning}
Prompting billion-parameter LLMs requires substantial compute. 
Finetuning smaller LLMs can be a more efficient alternative for scoring text items under different data constraints (table~\ref{tab:finetuning_results}).

\paragraph{If labeled data is available, finetuning offers a compelling alternative to prompting.} 
In \ImmFear, finetuning on as little as 500 pairs (with \llamai) can yield almost as good correlation and equal RMSE as the most performant prompting approach (5-shot pointwise prompting with \qwenbig). 
Overall, \deberta\ finetuned on 2000 pairwise examples is comparable or better than almost all pointwise prompting approaches on accuracy and correlation. 
In the \Grand\ dataset, finetuning \deberta\ with just 500 labeled pairs beats the best prompting approach, although it takes more data points to reach a comparable RMSE (\cref{fig:finetuning_train_size} in the appendix relates the number of training examples to performance).
In addition, finetuning \texttt{DeBERTa-v3-large} on pairwise data results in outputs that strike a balance between correlation and RMSE (\cref{fig:scatterplots_combined}).

Notably, finetuning a regression model on scalar outputs directly (here, BT estimates induced from the pairwise annotations in our datasets) does not yield the same benefits.
Finetuning regression models is less data-efficient and the relation between the number of training examples and RMSE tends to be unstable (see \cref{fig:finetuning_train_size_vs_regression}).

\begin{table}
\resizebox{0.8\textwidth}{!}{%
\begin{tabular}{llrr}
\toprule
 & & \multicolumn{2}{c}{Spearman's $\rho$} \\
 & & $\text{H}_\textsc{bt}$ & $\text{H}_\text{D}$ \\
\midrule
\multicolumn{3}{l}{\textsc{Immigration Fear}} \\
 & \emph{Human} &  \multicolumn{2}{c}{\emph{0.77}}\\
 & \llamasmall & 0.72 & 0.66 \\
 & \llamabig & 0.78 & 0.68 \\
 & \qwensmall & 0.75 & 0.66 \\
 & \qwenmid & 0.82 & 0.75 \\
 & \qwenbig & 0.85 & 0.72 \\
\multicolumn{3}{l}{\textsc{Ad-Negativity}} \\
 & \emph{Human} &  \multicolumn{2}{c}{\emph{0.85}}\\
 & \llamasmall & 0.88 & 0.88 \\
 & \llamabig & 0.89 & 0.91 \\
 & \qwensmall & 0.87 & 0.87 \\
 & \qwenmid & 0.88 & 0.89 \\
 & \qwenbig & 0.91 & 0.90 \\
\bottomrule
\end{tabular}
}
\caption{Human-to-human agreement is comparable to human-to-LLM agreement. The left column reports the correlation between LLM pointwise scores and Bradley-Terry scores induced from human pairwise ranks $\text{H}_\textsc{bt}$ (as elsewhere in the text); the right column is $\rho$ between LLM pointwise scores and \emph{direct} human annotations from experts, $\text{H}_\text{D}$. The \emph{human} row is the correlation between these two human annotation types, $\rho(\text{H}_\textsc{bt}, \text{H}_\textsc{D})$.}
\label{tab:direct-agreement}
\end{table}

\section{When do models and humans disagree?}
In this section, we include some additional findings to help contextualize our results.

\paragraph{Comparison with pointwise expert scoring.} \citet{carlson_pairwise_2017} compare their induced Bradley-Terry scores with \emph{direct} pointwise annotations from experts.
This measurement gives a rough sense of human-to-human agreement across different annotation methods; we can then compare this value to LLM-to-human scores.
Taking the \textsc{Ad} data as an example, \qwenbig{} with pointwise prompting has a Spearman's $\rho=0.91$ with the BT scores derived from human \emph{pairwise} annotations (per earlier results); the direct expert scores have $\rho=0.85$  with the same human BT scores (and 0.90 with the same direct LLM scores).
The findings from the other models and dataset (\cref{tab:direct-agreement}) suggest that LLM--human agreement between pointwise and pairwise is roughly on par with that of human--human.

\paragraph{Analysis of items with contrasting scorings.}

The pointwise LLM scores and the reference BT scores obtained from human pairwise annotations can disagree considerably in some instances (\cref{fig:scatterplots_combined}).
We examine such cases in the \textsc{ImmigrationFear} dataset to assess whether such scoring ``errors'' might be due to noise in BT scores obtained from human annotations (details in \Cref{apx:disagreement_pairs_analysis}).
Specifically, we select pairs of items for which the relative ranking induced from their LLM and BT scores contrast strongly.
We then distribute a sample of these pairs for a pairwise comparison annotation by four independent annotators (two of which are authors).
This analysis suggests that in the vast majority of these cases, the LLM score-based ranking is more aligned with our annotators' aggregate judgment (\cref{tab:disagreement_pairs_win_rates}).

\section{Conclusions}
The uptake of LLMs in social science research is high and increasing:\footnote{%
    On Scopus (\url{scopus.com}), ``large language model'' yields 1,500 social science articles in 2023 and 5,400 in 2024 (using \texttt{SUBJAREA `SOCI'}).
} %
For downstream inferences to be valid, it is important to use them appropriately.
Our survey on scalar construct measurement aims to guide practitioners, with findings that may translate outside social science constructs.

\section*{Limitations}

\paragraph{Dataset selection and generalization.}
While our datasets are diverse regarding the kind and complexity of social science constructs they cover, they obviously do not span the full breadth of possible scalar constructs.
Our study focuses on three social science constructs that represent (political) communication phenomena related to emotions, political attack behaviors, and rhetorical style.
However, it is possible that our results and the implications they have for applied researchers may not translate to other conceptually continuous concepts from other domains, like the level of anxiety expressed in user messages.
Moreover, our datasets only include English-language texts with applications focused on U.S. politics.
The generalization of our findings to other languages and country contexts is therefore an open question.
In such cases, it may be that the gap between prompted models and finetuning is larger, as it is less likely for the constructs to have been attested to in the training data.

\paragraph{Pair selection.} We predict ranks for the same pairs of items that were originally observed in the ground-truth data. Yet these would not exist in a real application: what sampling strategies for pairs are most effective? We consider that answering this question is likely the most fruitful direction for future work, potentially building on efforts in non-LLM settings \cite{Mikhailiuk2020pairwise,shima2022pairwise}.

For our part, we did undertake some preliminary investigations. In the finetuning setting, we tested whether there is a difference between training on a highly \emph{connected} graph compared to a disconnected one, holding the number of items equal. We didn’t find a significant (or even consistent) difference over datasets and models. But this indicates that any selection strategy may be serviceable, which is a potentially interesting finding in itself.

For pairwise prompting, using the \textsc{Immigration-Fear} data, we doubled the number of pairs for which we made predictions (in two iterations: adding new items and adding more links between items). There was little difference in the accuracy or correlation metrics in either case. That said, future work could evaluate other pair selection strategies, for example, by relying on semantic similarity or model confidence.

\paragraph{Exemplar selection.} Regarding our prompting methods, we note that the results we present focus on few-shot prompting with randomly selected exemplars.
We also studied more strategic exemplar selection, like choosing exemplars at anchoring points of the 1--9 scale.
These strategies did not consistently improve the scoring performance in our datasets, however.
Moreover, we did not examine the potential added value of Chain-of-Thought (CoT) prompting because \citet{wang2025improvingllmasajudgeinferencejudgment} convincingly demonstrated that it can collapse or otherwise disturb the judgment distribution and underlying token probabilities over response options in LLM-as-a-judge applications.

\paragraph{Computational considerations.} Further, deploying LLMs is compute-intensive, which may hinder practitioners' adoption of the methods we evaluate. 
We note, however, that our finding regarding the data efficiency and reliability of small finetuned models paves the way for valid text scoring with smaller, specialized models. 

\paragraph{Closed models.} While we cover several open model variants, we do not evaluate closed APIs, which often have a lower barrier to entry for non-technical users.
Closed APIs provide no, or only very limited, access to tokens' generation probabilities, preventing users from using this information for response calibration and debiasing.\footnote{
    Replicating our 5-shot pointwise scoring experiments with OpenAI's GPT-4o model showed that for typically more than 90\% of texts, at least one of the answer candidate tokens' probabilities was not among the top-20 most likely tokens' probabilities returned by the API.
}
Further, social science highly values reproducibility, which is at odds with the unpredictable updates and deprecation schedules of closed models \cite{barrie2024trust} (to the astute reader: yes, we did spin this limitation into a positive attribute).

\paragraph{Are human annotations a reasonable ground truth?} A crucial assumption underpinning the entire work is that human annotations are gold standard. Much prior work has challenged this paradigm \cite[e.g.,][]{clark-etal-2021-thats,hosking2024humanfeedback}: annotators can be biased, lack technical background, have insufficient context, make careless mistakes, or make other sorts of errors. How to validate measurement---either human or automated---is of course a major open question in the social sciences \cite{Adcock_Collier_2001}.

\section*{Acknowledgments} Work supported in part by
U.S. National Science Foundation award 2124270 and the ETH AI Center.

\bibliography{references.bib}

\clearpage
\appendix
\section{Methodological details}\label{apx:meth_details}

Below, we describe additional methodological details, covering the Bradley-Terry model and corrections for positional biases in LLM responses.

\subsection{The Bradley-Terry model}\label{apx:bradley-terry}

The Bradley–Terry model turns pairwise comparisons into a single latent scale.

Let the items (e.g., texts) in a set of pairwise comparisons be indexed by $i=1,\dots,n$.
Associate each item $i$ with a positive ``strength'' parameters $\alpha_i > 0$.
In our applications, these parameters estimate items' locations on the underlying (unobserved) latent dimension that represents the scalar construct under study (e.g., ad negativity).

Then, in a pairwise comparison of $i$ and $j$, the Bradley-Terry model models the probability that $i$ ``beats'' $j$ in terms of strength (equivalently, is ``higher'' on the underlying dimensions) as follows:
\[\Pr(i \succ j) = \frac{\alpha_i}{\alpha_i + \alpha_j}\]

Equivalently, writing $\alpha_i \equiv \exp(\lambda_i)$, with $\lambda_i \in \mathbb{R}$, the log odds of $i$ being chosen over $j$ is the following.
\[\text{logit}\left(\Pr(i \succ j) \right) = \log\left[ \frac{\Pr(i \succ j)}{\Pr(j \succ i)}\right] = \lambda_i - \lambda_j\]
$\lambda_i$ represents a latent propensity relevant to the comparison, such as ``ability,'' ``preference,'' etc., depending on the context.
The outcomes of pairwise comparisons are used to estimate these latent propensities relative to a chosen reference.
For identification, one can either choose a reference item or impose a constraint such as $\sum_{i} \lambda_i = 0$.

After estimation, the parameters may be rescaled (for example, to the unit interval) without affecting the fitted probabilities.
Probabilities are typically estimated using maximum likelihood, assuming independence of all pairwise comparisons.
Bias-reduced or penalized likelihood methods are often used to handle small samples or quasi- or complete separation. 

\subsection{Pairwise debiasing}\label{apx:pairwise_debiasing}

Our pairwise debiasing strategy borrows from~\citet{wang2025improvingllmasajudgeinferencejudgment} by prompting the model with different orderings of items in pairs to avoid positional biases \citep{zhao_calibrate_2021, han_prototypical_2023, wang_large_2023}.

Our pairwise prompts first explain the annotation task that define the criterion for comparison (e.g., emotional intensity) and then present a pair of text items in separate lines prefixed by ``Text 1'' and ``Text 2.''
The LLM is tasked to indicate which of the texts meets the comparison criterion and to respond either with ``1'' or ``2.''

\begin{table*}[!th]
    \centering
    \small
    \caption{Illustration of our pair augmentation strategy.}
    \label{tab:data_augmentation}
    {
    \begin{tabular}{@{}r c c l l@{}}\toprule
    & \multicolumn{2}{c}{ } & \multicolumn{2}{c}{position in prompt} \\\cmidrule(l{2pt}r{2pt}){2-3}\cmidrule(l{2pt}r{2pt}){4-5} 
    & swap label  & reverse order   &  \multicolumn{1}{c}{first option}        &  \multicolumn{1}{c}{second option}      \\\cmidrule(l{2pt}r{2pt}){1-5}\addlinespace
    \emph{original pair}   &    no          &      no          &  Text 1: ``item 1''  &  Text 2: ``item 2'' \\
    {augmentation 1} &    yes         &      no          &  Text 2: ``item 1''  &  Text 1: ``item 2'' \\
    {augmentation 2} &    no          &      yes         &  Text 1: ``item 2''  &  Text 2: ``item 1'' \\
    {augmentation 3} &    yes         &      yes         &  Text 2: ``item 2''  &  Text 1: ``item 1'' \\\bottomrule
    \end{tabular}
    }
\end{table*}%

Given this prompt format, our in-context learning approach consists of three steps: augmentation, prompting, and calibration.
We first augment our pair-level data by performing two augmentation operations on every pair illustrated in \Cref{tab:data_augmentation}.
In the prompting step, we then present one text pair at a time to the model, tasking it to respond with the label of the text item that is higher on the given comparison dimension (e.g., more emotionally intense).

Next, we generate and record the model's response, including its token probabilities for the two tokens ``1'' and ``2.''
\Cref{tab:fewshot_debiasing_illustration} for an example of an LLM's response and token probabilities for the original and augmentation versions of a text pair.

Finally, we debias the model's response as follows.
Let $p$ denote the probability for the token corresponding to label ``1'' in the model's generated response.
Let $i$ indicate whether the labels were swapped (augmentations 1 and 3), and $j$ whether text items' order was reversed (augmentations 2 and 3), where $i, j \in \{0, 1\}$ with $0$ indicating no swapping/reversing.
The model's preference score for the first text item in a pair is $p_{ij}$.
Specifically, the model's preference for the first item in a pair is captured by $p_{00}$, $1-p_{10}$, $p_{01}$, and $1-p_{11}$.
That is, given our augmentation strategy, which text item corresponds to choice ``1'' depends on whether the labels were swapped:
\[
y_{ij} = \begin{cases} 
    p_{ij}, & \text{if } i = 0 \text{ (i.e., \texttt{swapped})}\\
    1 - p_{ij}, & \text{if } i = 1 \text{ (i.e., not \texttt{swapped})}
\end{cases}
\]%
The \emph{debiased preference score} for the first text item, denoted $\hat{p}_{1}$, is the average of the \emph{aligned} preferences across the original pair and the three augmented conditions:
\[
\hat{p}_{1} = \frac{1}{4} \sum_{i=0}^{1} \sum_{j=0}^{1} y_{ij}
\]%
The corresponding debiased preference for the second item i $\hat{p}_{2} = 1 - \hat{p}_{1}$.
For example, for the pair shown in \Cref{tab:fewshot_debiasing_illustration}, the debiased preference scores are $\hat{p}_{1} = 0.688$ and $\hat{p}_{2} = 0.312$.
The model's debiased choice is therefore ``1''.

\begin{table*}[!ht]
    \centering
    \small
    \caption{%
        Illustration of LLM's preferences for response options and actual response (``choice'') for different augmentations of the same text pair (``original'').
    }
    \label{tab:fewshot_debiasing_illustration}
    {
    \begin{tabular}{@{}r *{5}{c}@{}}\toprule
        & \multicolumn{2}{c}{ } & \multicolumn{3}{c}{Model} \\ \cmidrule(l{2pt}r{2pt}){4-6}
       & \multicolumn{2}{c}{augmentations} & \multicolumn{2}{c}{token probs.} & \\ \cmidrule(l{2pt}r{2pt}){2-3} \cmidrule(l{2pt}r{2pt}){4-5}
 & {labels swapped} & {order reversed} & ``1'' & ``2'' & choice \\\midrule
        \emph{original pair}  &  no &  no & 0.996 & 0.004 & ``1'' \\
        augmentation 1 & yes &  no & 0.699 & 0.301 & ``1'' \\
        augmentation 3 &  no & yes & 0.197 & 0.803 & ``2'' \\
        augmentation 3 & yes & yes & 0.651 & 0.349 & ``1'' \\
        \bottomrule
    \end{tabular}
    }
\end{table*}

\section{Experiment details}

\subsection{GPU specification for LLM inference}\label{apx:gpu_specs}

We ran LLM inferences on local hardware and the ETH Zurich's GPU cluster EULER, depending on model size.
\Cref{tab:gpu_specs} provides details.

\begin{table}[h]
\caption{Overview of models, quantization, GPU hardware, and environment.}
\label{tab:gpu_specs}
\small
\centering
\resizebox{\textwidth}{!}%
{
\bgroup
\setlength\tabcolsep{3pt}
\begin{tabular}{@{}lrlr@{}}
\toprule
\textbf{Model} & \textbf{Quant.} & \textbf{GPU(s)} & \textbf{GPU RAM} \\
\midrule
\qwensmall   & 4-bit & 1 $\times$ NVIDIA GeForce RTX 4090 & 24,564 MB \\
\qwenmid  & 4-bit & 1 $\times$ NVIDIA A100-PCIE-40GB  & 40,960 MB   \\
              & \emph{none}  & 3 $\times$ NVIDIA GeForce RTX 4090  & 73,692 MB \\
\qwenbig & 4-bit & 3 $\times$ NVIDIA GeForce RTX 4090 &  73,692 MB  \\\addlinespace
\llamasmall   & 4-bit & 1 $\times$ NVIDIA GeForce RTX 4090 & 24,564 MB \\
\llamabig & 4-bit & 3 $\times$ NVIDIA GeForce RTX 4090 &  73,692 MB  \\
\bottomrule
\end{tabular}
\egroup
}
\end{table}%

\subsection{Finetuning hyperparameters}\label{apx:finetuning_hparams}

We use default hyperparameters for the finetuned models (\cref{tab:finetuning_hparams}), except for the learning rate, which we optimize on a validation set. Specifically, we sweep from the order of magnitude below the default to an order of magnitude above, in even steps, maximizing accuracy on a validation set from a distinct pairwise annotation dataset (\citealt{benoit_measuring_2019}, which we do not report on in the main text due to a large amount of annotation noise). The specific values are (\textbf{selected in bold}).
\begin{compactitem}
    \item \deberta: $6.00 \times 10^{-7}$, $1.29 \times 10^{-6}$, $2.78 \times 10^{-6}$, $6.00 \times 10^{-6}$, $1.29 \times 10^{-5}$, $\mathbf{2.78 \times 10^{-5}}$, $6.00 \times 10^{-5}$
    \item \llamasmall: $2.00 \times 10^{-5}$, $4.31 \times 10^{-5}$, $9.28 \times 10^{-5}$, $2.00 \times 10^{-4}$, $4.31 \times 10^{-4}$, $\mathbf{9.28 \times 10^{-4}}$, $2.00 \times 10^{-3}$
    \item \modbert: $1.42 \times 10^{-5}$, $2.53 \times 10^{-5}$, $4.50 \times 10^{-5}$, $\mathbf{8.00 \times 10^{-5}}$, $1.42 \times 10^{-4}$, $2.53 \times 10^{-4}$, $4.50 \times 10^{-4}$
\end{compactitem}
The maximum sequence length for all models is 384 tokens, which covers over 90\% of data points.

\begin{table*}[ht]
\caption{Finetuning model hyperparameters. -- Indicates the default in the Huggingface \texttt{transformers} library.}
\label{tab:finetuning_hparams}
\small
\centering
\begin{tabular}{@{}lrrr@{}}
\toprule
\textbf{Hyperparameter} & \modbert & \deberta & \llamasmall \\
\midrule
Learning rate & 8e-5 & 2.78e-5 & 9.28e-4 \\
Epochs & 5 & 10 & 5 \\
Batch size & 32 & 8 & 4 \\
Gradient accum. steps & 1 & 8 & 6 \\
Weight decay & 1e-5 & -- & 0.001 \\
Adam $\beta_1$ & 0.9 & -- & -- \\
Adam $\beta_2$ & 0.98 & -- & -- \\
Adam $\epsilon$ & 1e-6 & -- & -- \\
LoRA r & -- & -- & 8 \\
LoRA $\alpha$ & -- & -- & 32 \\
LoRA dropout & -- & -- & 0.1 \\
Precision & BF16 & FP32 & BF16 \\
Quantization & -- & -- & 8-bit \\
\bottomrule
\end{tabular}
\end{table*}%

\section{Evaluation of scoring methods through pairwise human comparison of text items with contrasting scores}\label{apx:disagreement_pairs_analysis}

In this analysis, we examine text items in the \ImmFear\ dataset that received very different scores on the underlying construct 
through human pairwise comparisons and LLM prompting. 
The benchmark in this analysis is a set of newly collected pairwise comparison decisions of four independent annotators.
The goal of this analysis is to assess which scoring method out of the following two yields scores that are more in line with these new annotators' judgments:
\begin{enumerate}
\tightlist
    \item The scores $\hat{\mathbf{s}}^{\text{BT}}$ obtained by fitting a Bradley-Terry (BT) model to the original singly-labeled human comparison judgments, \textit{or}, 
    \item The scores $\hat{\mathbf{s}}^{\text{LLM}}$ obtained through direct, pointwise LLM scoring with token probability-weighted averaging.
\end{enumerate}
We focus on instances of disagreement between these methods to better understand the limitations of each of these scoring methods.

Because this analysis focuses on discrepant cases and we expected pointwise scoring to be particularly difficult in such cases, we opted for a pairwise comparison to produce the relevant judgments data.
In particular, we sampled \emph{pairs of texts} for which the human comparison-based BT scores result in different pairwise rankings than the token probability-weighted LLM scores obtained through direct scoring.
That is, we focus on pairs of texts items $(i, j)$ for which $\hat{s}_{i}^\text{BT}> \hat{s}_{j}^\text{BT}$ but $\hat{s}_{i}^\text{LLM} < \hat{s}_{h}^\text{LLM}$, or \emph{vice versa}.
Our annotators then judged these pairs applying the original pairwise comparison construct \cite{carlson_pairwise_2017}.

Notably, the annotators were blind to which scoring approach yielded which relative ranking of the items.
This allows us to compute an unbiased win rate for the two methods.

\subsection{Sampling}

We based our analysis on the full set of texts in the test set of the \textsc{ImmigrationFear} dataset.
We use the scores of these items obtained by
    (a) fitting a Bradley-Terry model to human annotators (aggregated) pairwise comparison judgments and 
    (b) probability-weighted averaging of an LLM's direct scoring responses in 5-shot prompting with \qwenbig.
We denote these scores as $\hat{\mathbf{s}}^\text{BT}$ and $\hat{\mathbf{s}}^\text{LLM}$, respectively.

We rescaled each scoring variable to the range 0--1 because the BT scores are not on the 1--9 scale used for LLM prompting.
To avoid annotators comparing pairs of very long or very short texts, we subset the set of items based on the character counts of their texts to those within the 10\textsuperscript{th}--90\textsuperscript{th} percentile range, retaining $n$ items.

We then constructed the full schedule of $\frac{n\times(n-1)} {2}$ pairwise comparisons between these items.
For each pair of items $(i, j)$, used the scores $\hat{s}_{i}^\text{BT}, \hat{s}_{j}^\text{BT}$ and $\hat{s}_{i}^\text{LLM}, \hat{s}_{j}^\text{LLM}$ to induce pairwise comparisons for each scoring method.
This yields $c_{(i,j)}^\text{BT}, c_{(i,j)}^\text{LLM} \in {1, 2, 0}$ for each pair $(i,j)$, where $1$ ($2$) indicates that the first (second) item was chosen and $0$ indicates a tie.
Further, we compute 
$d_{(i,j)}^\text{BT} = \hat{s}_{i}^\text{BT} - \hat{s}_{j}^\text{BT}$ 
and 
$d_{(i,j)}^\text{LLM} = \hat{s}_{i}^\text{LLM} - \hat{s}_{j}^\text{LLM}$
to measure the strength and direction of disagreement between a pair's items for each scoring method.

We subset the pairwise comparison judgments to pairs in which items' texts are in the same text length decile to prevent text length from influencing annotators' judgments.
Further, to focus on clear-cut discrepancies, we keep only pairs for which $c_\text{BT} \neq c_\text{LLM}$ and none of $c_\text{BT}$ and $c_\text{LLM}$ indicate a tie.

\begin{figure}[t]
    \centering
    \includegraphics[width=\columnwidth]{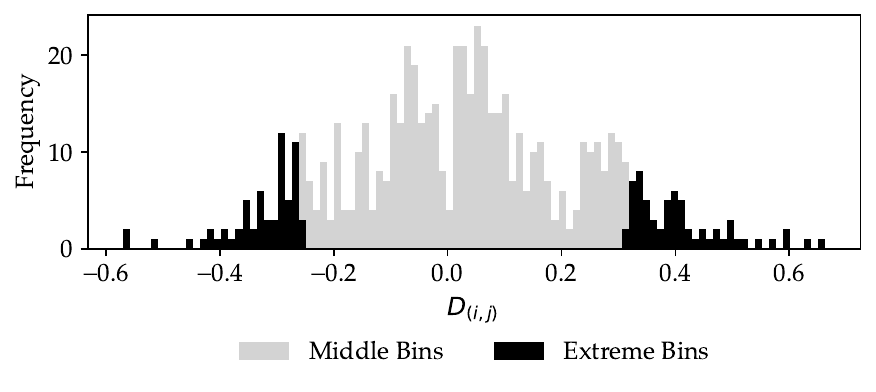}
    \caption{%
        Distribution of $D_{(i,j)}$ values in constructed pairwise comparisons.
        $D_{(i,j)}$ values measure the difference between the differences of a pair of items' BT and LLM-based scores, $d_{(i,j)}^\text{BT}$ and $d_{(i,j)}^\text{LLM}$, and measures the extend to (and direction in) which these two scoring methods' scores for the items in the pair disagree.
        Ranges of the histogram shaded in black indicate the set of pairs we sampled from for manual annotation.
    }
    \label{fig:sampled_disagreement_pairs_diffs_diff}
\end{figure}

We then compute $D_{(i,j)} = d_{(i,j)}^\text{BT} - d_{(i,j)}^\text{LLM}$ to obtain an indicator of the magnitude and direction of disagreement between the scoring methods' pairwise comparison score differences.
The distribution of these pairwise difference values is shown in \Cref{fig:sampled_disagreement_pairs_diffs_diff}.

In this subset, we grouped pairs based on $D$ into ten percentile bins.
We then sampled 30 pairs from the most extreme bins (shaded in black in \Cref{fig:sampled_disagreement_pairs_diffs_diff}) for blind and independent pairwise comparison through our annotators.
\Cref{tab:sampled_disagreement_pairs} shows four examples from this sample.

\begin{table*}[!t]
    \centering
    \small
    \caption{Examples of pairs of texts sampled from the \textsc{ImmigrationFear} dataset for scoring method evaluation. The item-level scores produced by the LLM (\qwenbig) and through fitting a BT model to human pairwise annotations, respectively, ought to measure the degree to which the text expresses fear, anxiety, or worry about the negative impact of immigration in the U.S.The pairs were sampled from the set of texts for which the pairwise comparison judgment induced from LLM scores and the human annotation-based BT scores disagree. Accordingly, the sampled cases focus on pairs of texts for which the different scoring methods result in contrasting pairwise ranking decisions.}
    \label{tab:sampled_disagreement_pairs}
    \begin{tabular}{L{5cm}L{5cm}cc}
        \toprule
        \multicolumn{2}{c}{Texts} & \multicolumn{2}{c}{Induced choice} \\\cmidrule(lr){1-2}\cmidrule(lr){3-4}\addlinespace
        Text A & Text B & $c_{(i,j)}^\text{BT}$ & $c_{(i,j)}^\text{LLM}$ \\
        \midrule
        i do not like the ifea of illegal immigration, but i also think it is too difficult to legally get residency in this country, especially if you are seeking asylum from religious persecution. & immigrants who have married an american but their spouses die are still sometimes kicked out of the country once they are widowed. i am also concerned about how difficult it is to get a green card. & Text A & Text B \\\addlinespace
        illegal immigration is a drain on the welfare system, although it seems to provide laborers for jobs americans don't want, legal immigration brings  technically skilled workers (h-1b), my great grandparents, and other people who can contribute to the country. & our immigration system doesn't work. beauacrats are sitting on their duffs and letting undocummented aliens, often subversive ones, flood our country. i don't see any efforts to repair the situation. & Text A & Text B \\\addlinespace
        if you mean illegal immigration, i'm afraid of who might be getting into this country in unsecured borders. & we need to get the upper hand on immigration, and treat everyone equally. we don't need to just start handing out licenses. and let's keep america's \#1 language english!!! & Text B & Text A \\\addlinespace
        people from other countries trying to come here to live or work and they don't always have the proper paper work & i think its a good thing bc the ones that live n the usa and is legal means we can now have more houses and job and dont have to worry bout them and problems they cause & Text B & Text A \\
        \bottomrule
    \end{tabular}
\end{table*}

\subsection{Annotation}

We randomized the order of the 60 pairs in this sample and distributed them for pairwise judgment to four independent annotators via a custom annotation interface implemented in survey software Qualtrix.
Two of the annotators were from the authors' team; the other two were trained research assistants.

The annotators were tasked with completing the pairwise comparisons task, indicating which of the two statements expresses more fear, anxiety, or worry about the negative impact of immigrants or immigration on America.
This is the original task and wording used by \citet{carlson_pairwise_2017}.
Importantly, the annotators were blinded to the pairwise comparison judgments induced from the BT and LLM scores and could thus not have been influenced in favor of any of the two scoring methods.

Considering that the two scoring methods we examine in this paper disagree heavily on the relative rankings of items chosen for this analysis, making judgments on their ordering based on the comparison construct is often not straightforward. 
This is reflected in the relatively modest chance-adjusted inter-coder agreement in our newly collected pairwise comparison annotations.\footnote{Krippendorff's $\alpha$ is 0.413}

We address this limitation by aggregating annotators' pair-level judgments with two established methods: majority voting and fitting a Dawid-Skene (DS) per-annotator model \cite{dawid_maximum_1979}.
Given that we have four annotations per pair and three label classes, majority voting requires that the annotations pass a relatively high bar to find a majority ``winner.''
The DS model, in turn, handles the problem of inter-coder disagreement by estimating annotator-specific ability parameters that allow attributing parts of pair-level disagreements to varying annotator-level error patterns.

\begin{table*}
    \centering
    \small
    \caption{Examples of pairs of texts with annotation disagreement and a relatively high posterior label entropy ($>0.1$). 
    \emph{Notes:}
    ``posterior label'' indicates the label class with the highest posterior probability according to the Dawid-Skene model fitted to the annotations. 
    ``entropy'' indicates the entropy of an item's posterior label probabilities, which is a measure of uncertainty in the label estimate. }
    \label{tab:disagreement_pairs_uncertain_instances}
    \begin{tabular}{L{4.8cm}L{4.8cm}ccc}
        \toprule
        \multicolumn{2}{c}{Texts} &  & \multicolumn{2}{c}{DS estimate} \\\cmidrule(lr){1-2}\cmidrule(lr){4-5}\addlinespace
        Text A & Text B & Annotations & posterior label & entropy \\
        \midrule
        we need to seal the borders. both north and south. fine all the employers that hire illegals and also those who rent to them.  if they can't find an job nor a place to live they will go home. & the only thing that makes me worry is the econany. immigration did not that happen yrars ago. i know all men are not equal. the rich get richer the poore get poorer. are we not all gods children? & [2, 0, 1, 1] & 1 & 0.443 \\\addlinespace
        i think its a good thing bc the ones that live n the usa and is legal means we can now have more houses and job and dont have to worry bout them and problems they cause & i believe we need to protect our borders.  we need to be sure people entering our country are doing so legally. & [2, 1, 2, 2] & 2 & 0.249 \\\addlinespace
        i do not like the ifea of illegal immigration, but i also think it is too difficult to legally get residency in this country, especially if you are seeking asylum from religious persecution. & immigrants who have married an american but their spouses die are still sometimes kicked out of the country once they are widowed. i am also concerned about how difficult it is to get a green card. & [2, 0, 1, 1] & 1 & 0.443 \\\addlinespace
        americans are not receptive to speakers of other than english and people who live south of the rio grande. immigrants are treated poorly by average citizens. & allour ancestors immigrated here. immigration should be done legally. we shouldn't subsidize illegal immigrants with government money. & [2, 0, 0, 2] & 0 & 0.177 \\\addlinespace
        losing good paying jobs to illegal immigrants. illegal immigrants not paying taxes. the government not doing enough to take care of illegal immigrants. & more than one side: being over-run, drugs, lower-end jobs not being filled, families tramatized, people wanting a better life & [2, 1, 0, 1] & 1 & 0.634 \\\addlinespace
        confused.  i think that everyone should be a citizen so that we receive money from their for their portion of taxes.  i also think that our country needs the immigrants more than we think as they will do almost any job, which many american's don't want to do. & immigrants who have married an american but their spouses die are still sometimes kicked out of the country once they are widowed. i am also concerned about how difficult it is to get a green card. & [2, 0, 1, 0] & 1 & 0.690 \\\addlinespace
        something this country was built with. something the american population against immigrants coming into the country need to educate themsleves more in. & legal entry into a country. that challenge of illegal entrants and how to eliminate them. bring me your tired, your poor.... & [2, 0, 2, 2] & 2 & 0.610 \\
        public safety, effect on the economy, whether or not they're entering legally (and if not how unfair and upsetting that is), that they're not willing to learn english & where i live immigration has over populated our city and it is no joke.  we have to many here taking over and getting aid. & [2, 0, 0, 0] & 0 & 0.293 \\
        \bottomrule
    \end{tabular}
\end{table*}

Majority voting resulted in 21 cases in which text 1 won, 21 cases in which text 2 won, 13 ties, and five invalid labels.
The DS model, in turn, yields 29 cases in which text 1 won, 17 in which text 2 won, and 14 ties.

However, the labels induced through majority voting and the DS model have moderately strong agreement.\footnote{
    Cohen's $\kappa$ is 0.753 in cases for which a valid majority voting label could be determined.
}.
Inter-annotator disagreements correspond to posterior label uncertainty in the DS estimates, as illustrated in \Cref{tab:disagreement_pairs_uncertain_instances}.

Notably, the DS per-annotator ``ability'' estimates for the two RA annotators were, on average, lower than those of the two authors (0.697 vs. 0.740), suggesting that the observed aggregate-level disagreement is partially explained by the former annotator group's lower reliability.

\subsection{Results}

We use the pair-level labels induced from our annotators' pairwise comparison judgments to evaluate the two scoring methods in cases where their scores lead to contrasting pairwise rankings.
Specifically, we compute how often the pairwise comparisons induced from LLM scores and BR scores agree with our annotators' aggregated judgments.
Recall that we only focus on pairs in which the pairwise comparisons induced from the two methods' scores disagreed.
This allows us to compute win rates in all non-tie cases our annotators have identified.

\begin{table}[!t]
    \centering
    \small
    \caption{Win rate of scoring methods in a sample of pairs of for which the scoring methods yield contrasting relative rankings of items. Scoring methods evaluated are pointwise scoring with LLM and token probability weighting (``LLM scores'') and scoring by a BT model fitted to human pairwise comparisons annotations (``BT scores''). Values reported are the number of cases (and share) of pairs for which the pairwise judgments aggregated from four independent annotators align with the pairwise ranking of the respective scoring method. Rows report results for two annotations aggregation methods: majority voting and a Dawid-Skene per-annotator model.}
    \label{tab:disagreement_pairs_win_rates}
    \begin{tabular}{lcc}
    \toprule
    winner & Majority vote & Dawid-Skene model \\
    \midrule
    LLM scores & 41 (68\%) & 35 (69\%) \\
    \emph{neither} & 18 (30\%) & 15 (29\%) \\
    BT scores & 1 (2\%) & 1 (2\%) \\
    \bottomrule
    \end{tabular}
\end{table}

\Cref{tab:disagreement_pairs_win_rates} shows that in pairs of texts for which the relative pairwise ranking induced from human annotation-based BT scores and the LLM disagrees, the LLM scoring approach typically yields decisions that are more aligned with the blind judgments of our four independent annotators.
While our annotators declare some of the pairs as ties where both scoring methods (by case selection) declare a clear winner, the LLM scoring approach is the winner in all but one of the remaining cases, independent of the judgment aggregation method.
This suggests that the LLM scoring method is more aligned with our independent judgments of the underlying construct than the original human annotations.%

\section{Additional Results}\label{apx:additional_results}

\begin{figure*}[th]
    \centering
    \includegraphics[width=\columnwidth]{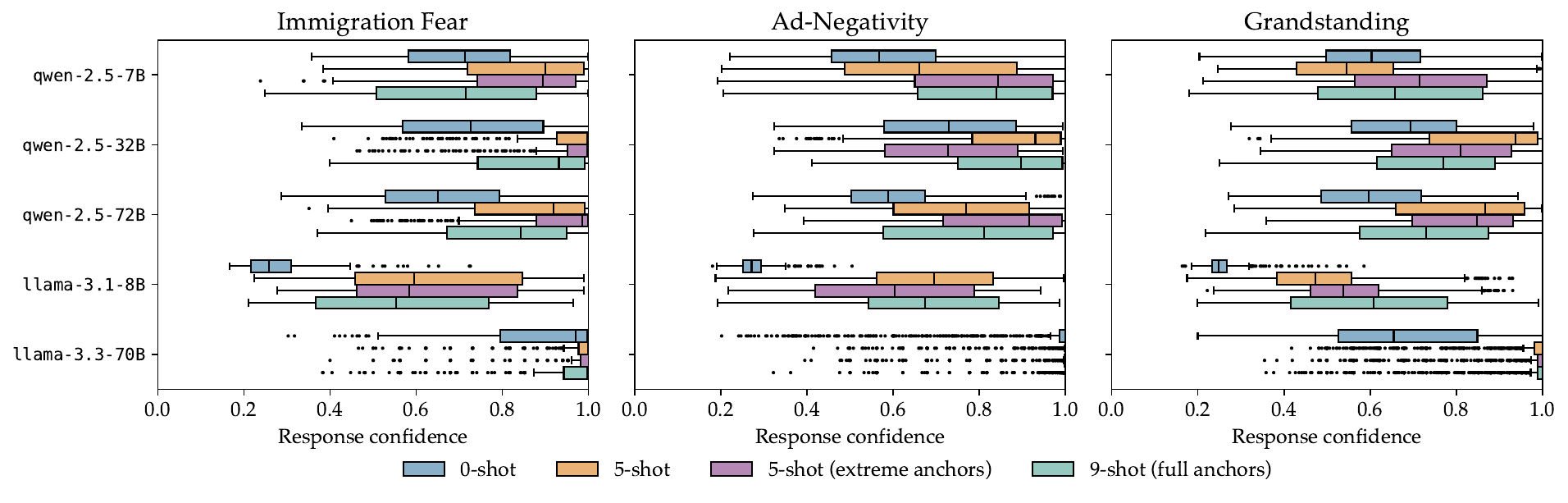}%
    \caption{LLM confidence in the modal response varies widely over model variants and prompting strategy, leading to inconsistent behavior that can impact downstream conclusions.}
    \label{fig:prompting_response_confidence_distribution}
\end{figure*}

\Cref{fig:prompting_response_confidence_distribution} shows the distribution of LLM confidences in the most-probable (modal) response over multiple model types and prompting strategies.
Any given design decision can have a large impact on confidence.

\Cref{tab:gpt4o_results} compares the pointwise scoring performance of the two open-weights models we use to the performance of GPT-4o based on the Ad-Negativity data.
\Cref{tab:no_quant_prompting_results} shows that our prompting results for pointwise scoring with token probability weighting are robust to running inference without 4-bit quantization.

\begin{table*}[ht]
\caption{Comparison of pointwise scoring prompting results between open-weights and proprietary models in Ad-Negativity data. \emph{Notes:} Metrics reported are the pair classification accuracy (Acc) and scoring performance relative to ground-truth BT scores in terms of Spearman's rank correlation ($\rho$) and the root mean squared error (RMSE, on scale 0–1). Values report are averages $\pm$ one standard deviation computed  by summarizing results across five folds.}
\label{tab:gpt4o_results}
\small
\bgroup
\setlength\tabcolsep{3pt}
\begin{tabular}{@{}>{\bfseries\ttfamily}rcccccc@{}}
\toprule
 & \multicolumn{3}{c}{zero-shot} & \multicolumn{3}{c}{5-shot} \\\cmidrule(lr){2-4}\cmidrule(lr){5-7}
 & Acc & $\rho$ & RMSE & Acc & $\rho$ & RMSE \\
\midrule
GPT-4o & 0.786±0.011 & 0.889±0.016 & 0.253±0.009 & 0.802±0.010 & 0.917±0.010 & 0.209±0.010 \\
Llama-3.3-70b & 0.795±0.011 & 0.888±0.014 & 0.288±0.008 & 0.800±0.010 & 0.903±0.017 & 0.250±0.010 \\
Qwen-2.5-72b & 0.806±0.010 & 0.906±0.011 & 0.134±0.007 & 0.807±0.012 & 0.918±0.010 & 0.165±0.011 \\
\bottomrule
\end{tabular}
\egroup
\end{table*}

\begin{table*}[ht]
\caption{Comparison of prompting results for 5-shot pointwise scoring with and without 4-bit quantization for selected datasets and models. \emph{Notes:} Rows correspond to the model (e.g., qwen/llama) and few-shot method (0-shot or 5-shot). Metrics reported are the pair classification accuracy (Acc) and scoring performance relative to ground-truth BT scores in terms of Spearman's rank correlation ($\rho$) and the root mean squared error (RMSE, on scale 0–1). Values report averages $\pm$ one standard deviation computed based on 25 bootstrapped estimates in test split. Values in bold mark the best result for a dataset and metric and values underlined mark results within one standard error of the best result. }
\label{tab:no_quant_prompting_results}
\small
\bgroup
\setlength\tabcolsep{3pt}
\begin{tabular}{@{}rlrccc@{}}
\toprule
 &  &  & Acc & $\rho$ & RMSE \\
Dataset & Model & Quantization &  &  &  \\
\midrule
\textsc{Immigration Fear} & \textbf{\texttt{Qwen-2.5-32b}} & 4-bit & 0.768±0.007 & 0.853±0.031 & 0.222±0.011 \\
 &  & \emph{none} & 0.763±0.007 & 0.853±0.034 & 0.232±0.012 \\
\cmidrule(lr){2-6}
 & \textbf{\texttt{llama-3.1-8b}} & 4-bit & 0.740±0.008 & 0.780±0.037 & 0.266±0.011 \\
 &  & \emph{none} & 0.748±0.008 & 0.795±0.034 & 0.259±0.012 \\
\cmidrule(lr){1-6} \cmidrule(lr){2-6}
\textsc{Ad-Negativity} & \textbf{\texttt{Qwen-2.5-32b}} & 4-bit & 0.799±0.010 & 0.896±0.016 & 0.158±0.008 \\
 &  & \emph{none} & 0.789±0.010 & 0.897±0.016 & 0.163±0.009 \\
\cmidrule(lr){2-6}
 & \textbf{\texttt{Llama-3.1-8b}} & 4-bit & 0.806±0.009 & 0.899±0.014 & 0.196±0.007 \\
 &  & \emph{none} & 0.794±0.009 & 0.897±0.014 & 0.222±0.008 \\
\bottomrule
\end{tabular}
\egroup
\end{table*}

\begin{figure}[!h]
    \centering
    \includegraphics[width=\textwidth]{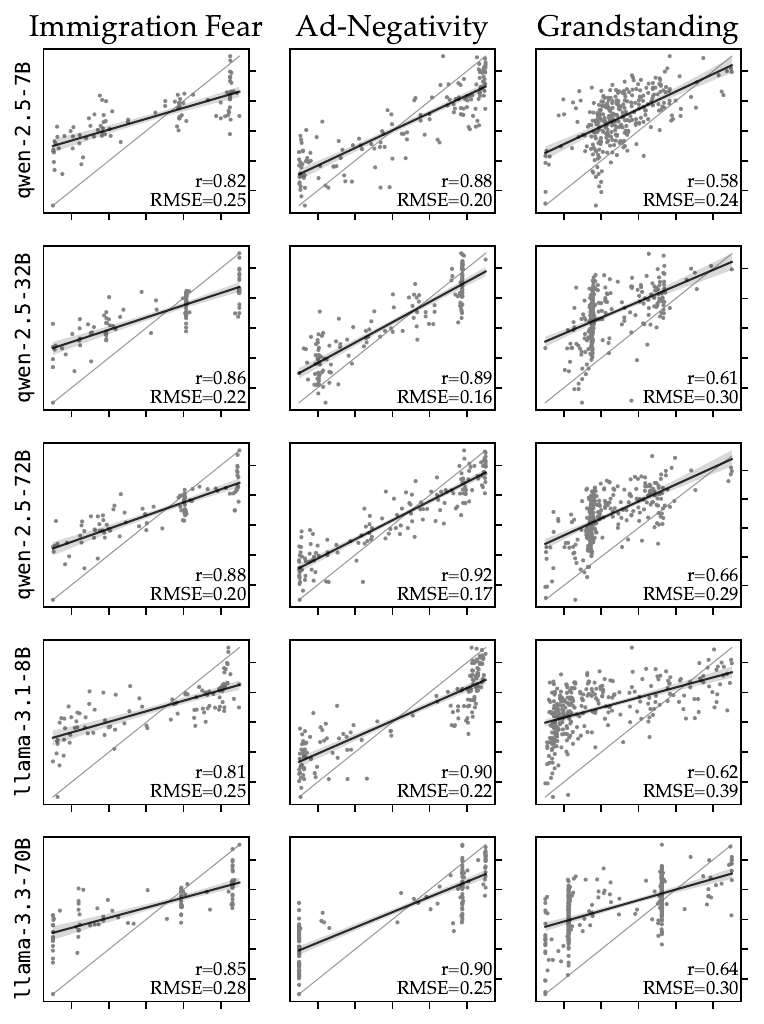}%
    \caption{
        Relation between model responses and ground-truth BT scores in prompted models' probability-weighted average pointwise scores (5-shot).
    }
    \label{fig:prompting_averags_scatterplots_5shot}
\end{figure}

\begin{figure}[!h]
    \centering
    \includegraphics[width=\textwidth]{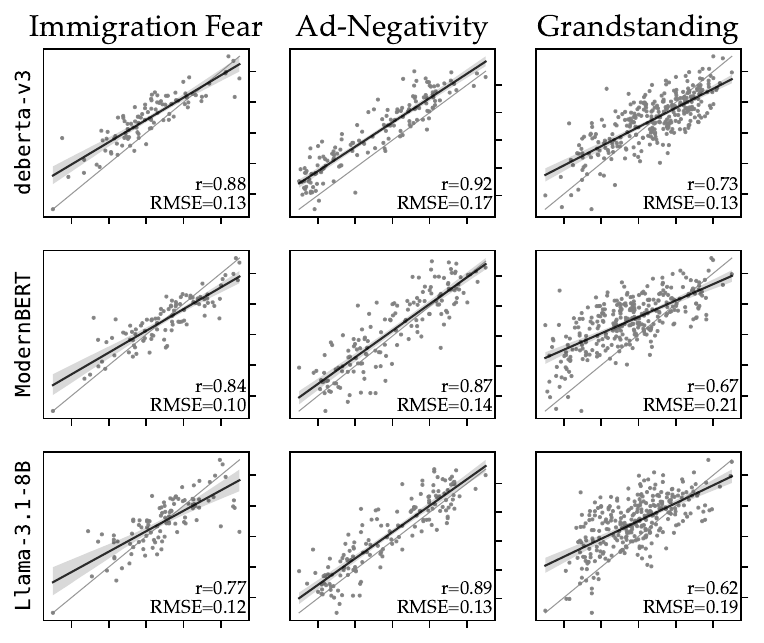}%
    \caption{
        Relation between model responses and ground-truth BT scores in models fine-tuned on all examples in datasets' training splits (
            2,729 in the \ImmFear,
            6,760 in the \AdNeg, 
            and
            32,308 in the \ImmFear data). 
    }
    \label{fig:finetuning_scores_scatterplots_nall}
\end{figure}

Complementing the selected results highlighted in \Cref{fig:scatterplots_combined},
\Cref{fig:prompting_averags_scatterplots_5shot} shows the relation between model responses and ground-truth BT scores in prompted models' probability-weighted average pointwise scores (5-shot), and
\Cref{fig:finetuning_scores_scatterplots_nall} the relation between predicted scores and ground-truth BT scores in models fine-tuned on all examples in datasets' training splits.

\begin{figure*}[!h]
    \centering
    \includegraphics[width=\textwidth]{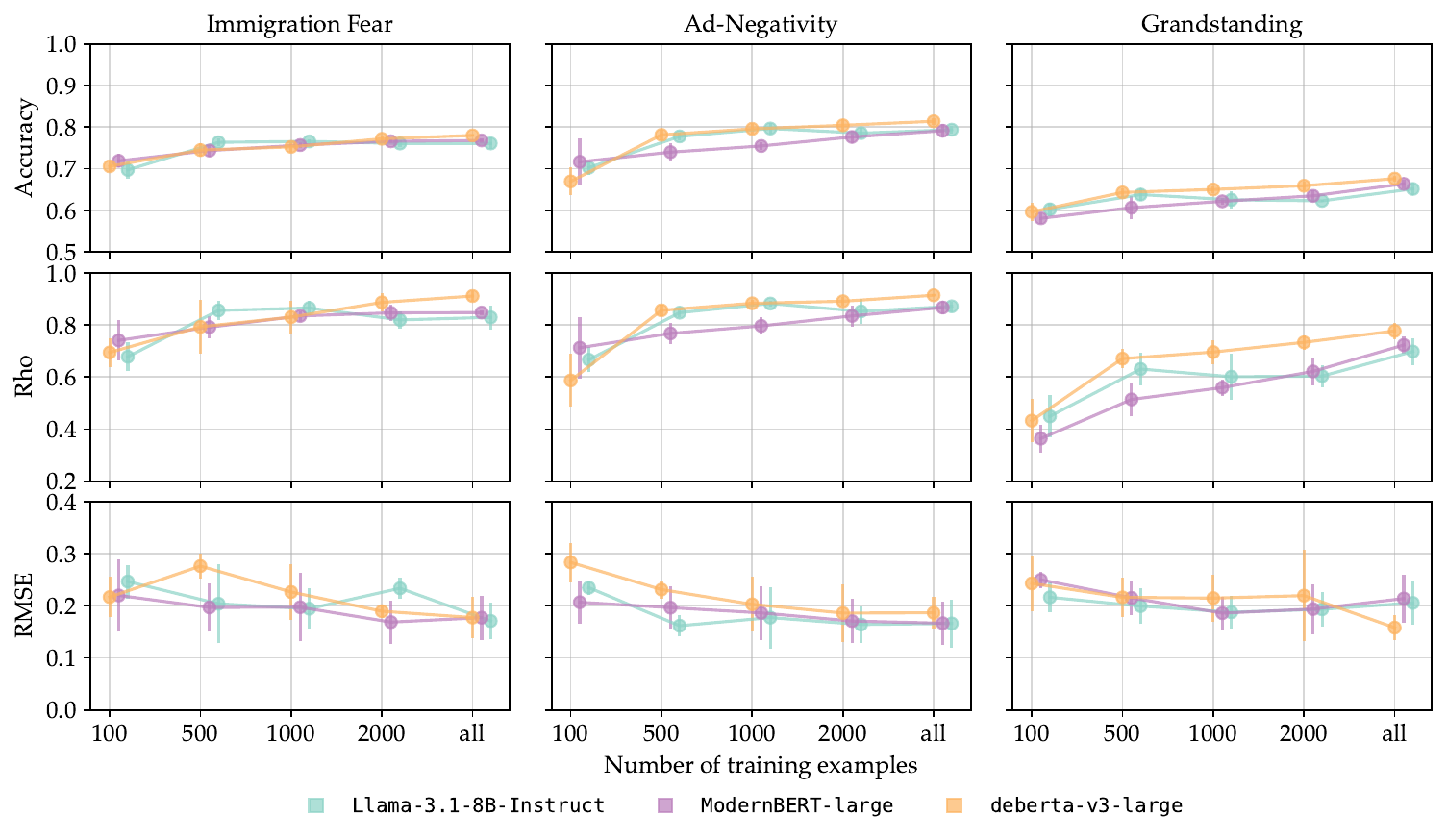}%
    \caption{
        Classification and scoring performance of finetuned models as a function of the number of training examples.
        ``all'' refers to all pairs in the training split (
            2,729 in the \ImmFear,
            6,760 in the \AdNeg, 
            and
            32,308 in the \ImmFear data).
        Points and vertical bars report averages ± one standard deviation computed by summarizing results across five folds.
    }
    \label{fig:finetuning_train_size}
\end{figure*}

\begin{figure*}[!h]
    \centering
    \includegraphics[width=\textwidth]{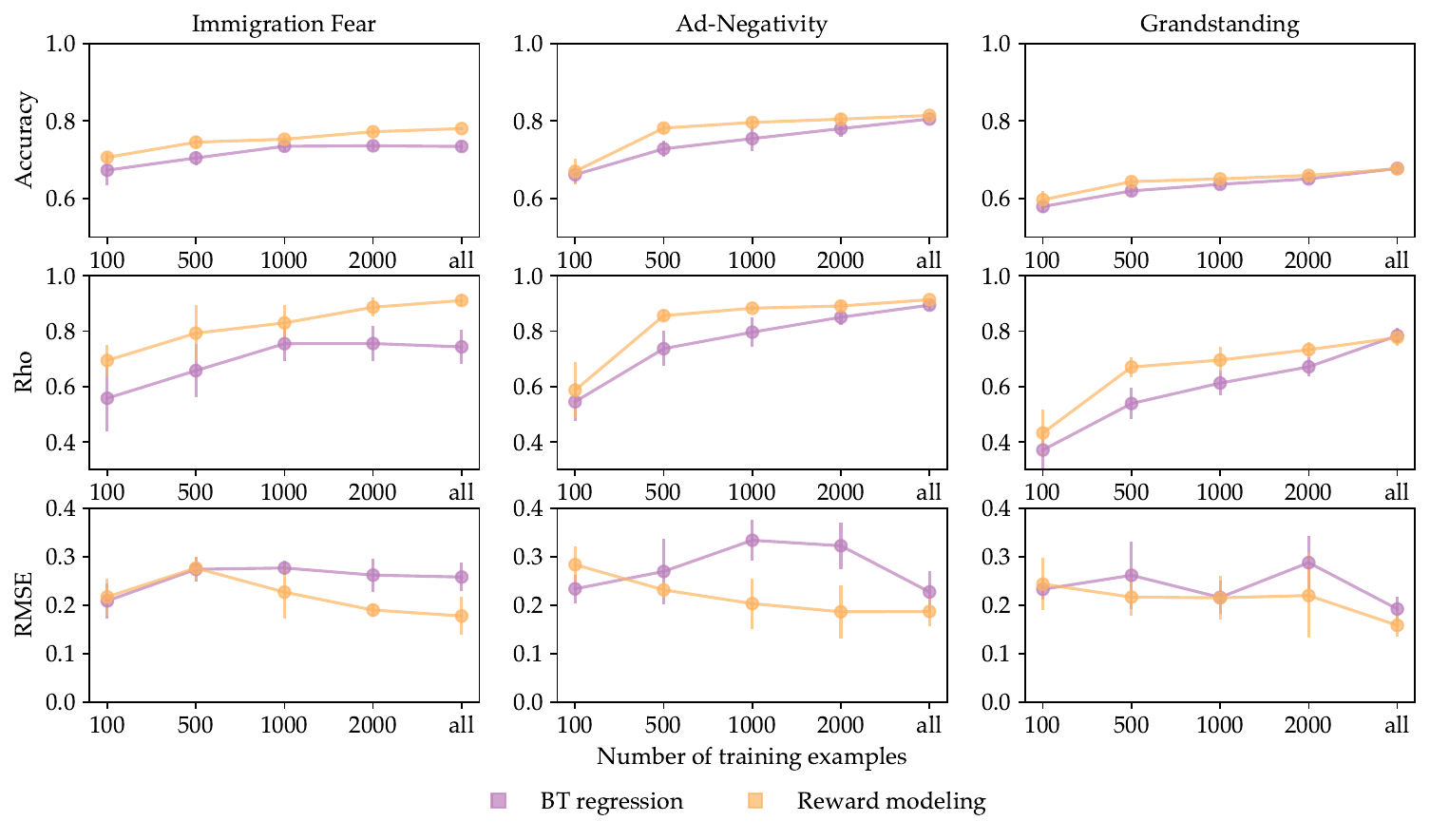}%
    \caption{
        Classification and scoring performance of finetuned reward and regression models as a function of the number of training examples.
        Results shown for finetunes of \texttt{deberta-v3-large}.
        ``all'' refers to all pairs in the training split (%
            2,729 in the \ImmFear,
            6,760 in the \AdNeg, 
            and
            32,308 in the \ImmFear data).
        Points and vertical bars report averages ± one standard deviation computed by summarizing results across five folds.
    }
    \label{fig:finetuning_train_size_vs_regression}
\end{figure*}

\Cref{fig:finetuning_train_size}, in turn, shows how finetuned models' classification and scoring performance changes with the number of training examples used for finetuning.
\Cref{fig:finetuning_train_size_vs_regression} contrasts finetuned reward models' performance with those of regression models finetuned using inferred BT scores instead of pairwise label data.

\section{Prompts}\label{apx:prompts}

\begin{prompt}[title={Pairwise comparison prompt used for \ImmFear\ task in Carlson and Montgomery (2017) data.}, label=prompt:montogmery2017_immigration_pairwise]
\begin{lstlisting}
Your tasks is to **indicate which of these two statements expresses more fear, anxiety, or worry about the negative impact of immigrants or immigration on America.** 

Note that we are not interested in whether the writer dislikes immigrants, wants them to go home, resents them, or blames them. We are only interested in whether the writer is expressing fear, anxiety, or worry.

## Input

{text}

**Response**: [Output only "1" or "2"]
\end{lstlisting}
\end{prompt}

\begin{prompt}[title={Absolute scoring prompt used for \ImmFear\ task in Carlson and Montgomery (2017) data.}, label=prompt:montogmery2017_immigration_itemwise]
\begin{lstlisting}
## Task

Your tasks is to **rate the statement below in terms of the level of fear, anxiety, or worry about the negative impact of immigrants or immigration on America expressed in it on a scale from 1 (none) to 9 (extreme).**

Note that we are not interested in whether the writer dislikes immigrants, wants them to go home, resents them, or blames them. We are only interested in whether the writer is expressing fear, anxiety, or worry.

## Input

{text}

**Response**: [Output only a score between 1 and 9]
\end{lstlisting}
\end{prompt}

\begin{prompt}[title={Pairwise comparison prompt used for \AdNeg\ task in Carlson and Montgomery (2017) data.}, label=prompt:montogmery2017_wiscads_pairwise]
\begin{lstlisting}
## Overview

This task involves reading the text of two television advertisements aired during the 2008 U.S. Senate elections. Each advertisement consists of about one paragraph of text. Researchers will use your responses to better understand the "tone'' of each political ad. 

You will see text from **two advertisements**. Your job is to read both and select the one that is:
 
- **most** _negative_ towards the candidate(s) mentioned, or;
- **least** _positive_ about the candidate(s) mentioned. 

Some of these choices will be very clear, but others will require you to use your best judgement.

## Details 

Here are a few rules of thumb to guide you:

- Ads that attack a candidate's personal characteristics (e.g., "Bob is dishonest.") are generally more negative than ads that attack a candidate's record or job performance (e.g., "Bob is too liberal.'').
- Ads that attack a specific candidate alone (e.g., "Bob is unqualified") are generally more negative than ads that contrast two candidates (e.g., "Bob is unqualified, but Jill is very experienced").
- Ads that attack a named individual (e.g., "Bob spent his time in Washington working for fat cats") are generally more negative than ads that attack a general group (e.g., "We need to stop those fat cats in Washington").
- Ads that state a policy position (e.g., "Bob will find everyone jobs") are generally less positive than ads that praise a candidate as a person (e.g., "Bob is a leader.").
- If both advertisements attack a candidate, pick whichever of the two advertisements is _**most** negative_.
- If both advertisements praise a candidate, pick whichever of the two advertisements is _**least** positive_.
- Do not allow your own political opinions to influence your decisions. Your goal is to select the ad that other coders would recognize as the most negative (or least positive).

It is critical that you read each statement carefully. Skimming or reading quickly will result in low-quality evaluations.

## Background

The texts you will be reading were collected by the Wisconsin Advertising Project (WiscAds), which studies political advertisements in the United States. In this task, we are interested in ads from the the 2008 U.S. Senate elections.

WiscAds takes each television ad and creates a "storyboard" composed of the words included in the ad's voiceover. 

Here are a couple of things to remember:

1. Words in brackets (e.g., [Roberts]) indicate who is speaking. So, when the children speak, it looks like this: '[Kids]: "Right Pat!"'
2. The final line of each ad will always include a bracket [PFB], which is short for "Paid for By." In this case, the ad was paid for by the organization, "Pat Roberts for U.S. Senate." So, in this case, the last line will be: "[PFB:] Pat Roberts for U.S. Senate."
3. The ads you code will not include images. That means you will have to use only the text from the storyboard to code the ads.

## Task

Please read the two advertisement texts below. Your job is to read both and select the ad that is:

- **most** _negative_ towards the candidate(s) mentioned, or;
- **least** _positive_ about the candidate(s) mentioned. 

## Input

{text}

**Response**: [Output only "1" or "2"]
\end{lstlisting}
\end{prompt}

\begin{prompt}[title={Absolute scoring prompt used for \AdNeg\ task in Carlson and Montgomery (2017) data.}, label=prompt:montogmery2017_wiscads_itemwise]
\begin{lstlisting}
## Overview

This task involves reading the text of a television advertisement aired during the 2008 U.S. Senate elections. The advertisement consists of about one paragraph of text. Researchers will use your responses to better understand the "tone'' of the political ad. 

You will see the text of **an advertisement**. Your job is to read it and rate it on a scale ranging from 1 to 9 in terms of how:
 
- 9: _negative_ it is towards the candidate(s) mentioned, or;
- 1: _positive_ it is about the candidate(s) mentioned. 

Some of these choices will be very clear, but others will require you to use your best judgement.

## Details 

Here are a few rules of thumb to guide you:

- Ads that attack a candidate's personal characteristics (e.g., "Bob is dishonest.") are generally more negative than ads that attack a candidate's record or job performance (e.g., "Bob is too liberal.'').
- Ads that attack a specific candidate alone (e.g., "Bob is unqualified") are generally more negative than ads that contrast two candidates (e.g., "Bob is unqualified, but Jill is very experienced").
- Ads that attack a named individual (e.g., "Bob spent his time in Washington working for fat cats") are generally more negative than ads that attack a general group (e.g., "We need to stop those fat cats in Washington").
- Ads that state a policy position (e.g., "Bob will find everyone jobs") are generally less positive than ads that praise a candidate as a person (e.g., "Bob is a leader.").
- If both advertisements attack a candidate, pick whichever of the two advertisements is _**most** negative_.
- If both advertisements praise a candidate, pick whichever of the two advertisements is _**least** positive_.
- Do not allow your own political opinions to influence your decisions. Your goal is to select the ad that other coders would recognize as the most negative (or least positive).

It is critical that you read the statement carefully. Skimming or reading quickly will result in low-quality evaluations.

## Background

The text you will be reading were collected by the Wisconsin Advertising Project (WiscAds), which studies political advertisements in the United States. In this task, we are interested in ads from the the 2008 U.S. Senate elections.

WiscAds takes each television ad and creates a "storyboard" composed of the words included in the ad's voiceover. 

Here are a couple of things to remember:

1. Words in brackets (e.g., [Roberts]) indicate who is speaking. So, when the children speak, it looks like this: '[Kids]: "Right Pat!"'
2. The final line of each ad will always include a bracket [PFB], which is short for "Paid for By." In this case, the ad was paid for by the organization, "Pat Roberts for U.S. Senate." So, in this case, the last line will be: "[PFB:] Pat Roberts for U.S. Senate."
3. The ads you code will not include images. That means you will have to use only the text from the storyboard to code the ads.

## Task

Please read the advertisement text below. Your job is to read it and rate it on a scale ranging from 1 to 9 in terms of how: 

- 9: _negative_ it is towards the candidate(s) mentioned, or; 
- 1: _positive_ it is about the candidate(s) mentioned.

## Input

{text}

**Response**: [Output only a score between 1 and 9]
\end{lstlisting}
\end{prompt}

\begin{prompt}[title={Pairwise comparison prompt used for \Grand\ task in Park (2021) data.}, label=prompt:park2021_pairwise]
\begin{lstlisting}
## Overview

You will be presented two paragraphs from the House representatives' speeches during congressional hearings. Your task is to choose the paragraph that is relatively more opinionized/grandstanding or less factual/information-seeking.


## Background

To give you some background knowledge, congressional committees hold hearings for various purposes: to monitor executive branches, to collect information for legislations, to approve government nominees or budgeting plans, etc. A congressional hearing proceeds as follows: It starts with the committee chair's opening speech followed by other committee members' and witnesses' opening speeches. Then, the chair proceeds to a Q&A session where committee members ask questions to witnesses. Long speeches are broken down to paragraphs. Thus, some paragraphs you will compare can be part of a longer speech.


## Details

A speech is an **opinionized or grandstanding** speech if it does one of the following: 

1. Denouncing (or Praising) a person or an institution (e.g. a party, its members, president, a government agency, a witness or others)
2. Taking positions on a policy by approving or disapproving it (which includes subjective interpretation of a policy-relevant situation)
3. Asking questions just to embarrass or attack a witness:

A speech is a **factual or information-seeking** speech if it is one of the following:
   
1. Objective description of a policy-relevant situation
2. Asking witnesses questions for fact-checking or expert opinion-seeking

A speech is **neither opinionized nor information-seeking** if it falls into the following:

1. Procedural remarks: 
2. None of these mentioned above (No content):


## Important notes

- Consider that speeches can be placed onto a continuum of which one extreme end is opinionized/grandstanding speeches and the other extreme end is factual/information-seeking speeches. In the middle of the two ends, speeches that are neither the two including procedural speeches can be located.
- It is important that you read each speech extract carefully, and that you judge each by the standards listed above and the information in the text. 
- In comparing the two paragraphs, DO NOT make your judgments on your own knowledge of a person or a policy in question or on definitions of opinions different to those listed above. 
- Note that not all questions are information-seeking but can be part of grandstanding depending on what is being asked and how. Also, note that the length of a speech excerpt is irrelevant to and does not cue the type of speech.


## Task

Please read the two statements below and select the statement that is relatively more opinionized/grandstanding or less factual/information seeking.

- A statement is more opinionized or grandstanding if it denounces or praises an institution or a person, or expresses subjective views on a policy or a situation more explicitly and strongly.
- A statement is factual or information seeking if it gives objective description of a situation or asking witnesses for information or their opinion. 

Which of the two statements below is more opinionized/grandstanding or less factual/information seeking? 


{text}

**Response**: [Output only "1" or "2"]
\end{lstlisting}
\end{prompt}

\begin{prompt}[title={Absolute scoring prompt used for \Grand\ task in Park (2021) data.}, label=prompt:park2021_itemwise]
\begin{lstlisting}
## Overview

You will be presented a paragraph from the House representatives' speeches during congressional hearings. Your task is rate the paragraph on scale ranging from opinionized/grandstanding on one end to factual/information-seeking on the other.


## Background

To give you some background knowledge, congressional committees hold hearings for various purposes: to monitor executive branches, to collect information for legislations, to approve government nominees or budgeting plans, etc. A congressional hearing proceeds as follows: It starts with the committee chair's opening speech followed by other committee members' and witnesses' opening speeches. Then, the chair proceeds to a Q&A session where committee members ask questions to witnesses. Long speeches are broken down to paragraphs. Thus, the paragraphs you will rate can be part of a longer speech.


## Details

A speech is an **opinionized or grandstanding** speech if it does one of the following: 

1. Denouncing (or Praising) a person or an institution (e.g. a party, its members, president, a government agency, a witness or others)
2. Taking positions on a policy by approving or disapproving it (which includes subjective interpretation of a policy-relevant situation)
3. Asking questions just to embarrass or attack a witness

A speech is a **factual or information-seeking** speech if it is one of the following:
   
1. Objective description of a policy-relevant situation
2. Asking witnesses questions for fact-checking or expert opinion-seeking
     
A speech is **neither opinionized nor information-seeking** if it falls into the following:

1. Procedural remarks: 
2. None of these mentioned above (No content):


## Important notes

- Consider that speeches can be placed onto a continuum of which one extreme end is opinionized/grandstanding speeches and the other extreme end is factual/information-seeking speeches. In the middle of the two ends, speeches that are neither the two including procedural speeches can be located.
- It is important that you read the speech extract carefully, and that you judge it by the standards listed above and the information in the text. 
- In rating the paragraph, DO NOT make your judgment on your own knowledge of a person or a policy in question or on definitions of opinions different to those listed above. 
- Note that not all questions are information-seeking but can be part of grandstanding depending on what is being asked and how. Also, note that the length of a speech excerpt is irrelevant to and does not cue the type of speech.


## Task

Please read the statement and rate it on a scale ranging from 1 to 9 how opinionized/grandstanding (9) or factual/information seeking (1).

- A statement is **opinionized or grandstanding** if it denounces or praises an institution or a person, or expresses subjective views on a policy or a situation more explicitly and strongly.
- A statement is **factual or information seeking** if it gives objective description of a situation or asking witnesses for information or their opinion. 

On a scale ranging from 1 to 9, how opinionized/grandstanding (9) or factual/information seeking (1) is the statement shown below?

{text}

**Response**: [Output only a score between 1 and 9]
\end{lstlisting}
\end{prompt}

\end{document}